\documentclass[letterpaper]{article} 
\usepackage{aaai25}  
\usepackage{times}  
\usepackage{helvet}  
\usepackage{courier}  
\usepackage[hyphens]{url}  
\usepackage{graphicx} 
\usepackage{natbib}  
\usepackage{caption} 
\usepackage{algorithm}
\usepackage{algorithmic}

\usepackage{newfloat}
\usepackage{listings}
\DeclareCaptionStyle{ruled}{labelfont=normalfont,labelsep=colon,strut=off} 
\floatstyle{ruled}
\newfloat{listing}{tb}{lst}{}
\floatname{listing}{Listing}
\title{VG-TVP: Multimodal Procedural Planning via Visually Grounded Text-Video Prompting}
\author{
    Muhammet Furkan Ilaslan\textsuperscript{\rm 1, 2}\thanks{\ \ Corresponding author.},
    Ali Koksal\textsuperscript{\rm 2},
    Kevin Qinhong Lin\textsuperscript{\rm 1},
    Burak Satar\textsuperscript{\rm 2},\\
    Mike Zheng Shou\textsuperscript{\rm 1},
    Qianli Xu\textsuperscript{\rm 2}
}
\affiliations{
    \textsuperscript{\rm 1}Show Lab, National University of Singapore, Singapore\\
    \textsuperscript{\rm 2}Institute for Infocomm Research, Agency for Science, Technology, and Research (A*STAR), Singapore\\

    \{m.furkanilaslan, qinghonglin\}@u.nus.edu, mike.zheng.shou@gmail.com\\
    \{stuimf, koksal\_ali, burak\_satar, qxu\}@i2r.a-star.edu.sg\\
}

\usepackage{bibentry}

\begin{document}

\maketitle

\begin{abstract}
Large Language Model (LLM)-based agents have shown promise in procedural tasks, but the potential of multimodal instructions augmented by texts and videos to assist users remains under-explored. To address this gap, we propose the Visually Grounded Text-Video Prompting (VG-TVP) method which is a novel LLM-empowered Multimodal Procedural Planning (MPP) framework. It generates cohesive text and video procedural plans given a specified high-level objective. The main challenges are achieving textual and visual informativeness, temporal coherence, and accuracy in procedural plans. VG-TVP leverages the zero-shot reasoning capability of LLMs, the video-to-text generation ability of the video captioning models, and the text-to-video generation ability of diffusion models. VG-TVP improves the interaction between modalities by proposing a novel Fusion of Captioning (FoC) method and using Text-to-Video Bridge (T2V-B) and Video-to-Text Bridge (V2T-B). They allow LLMs to guide the generation of visually-grounded text plans and textual-grounded video plans. To address the scarcity of datasets suitable for MPP, we have curated a new dataset called Daily-Life Task Procedural Plans (Daily-PP). We conduct comprehensive experiments and benchmarks to evaluate human preferences (regarding textual and visual informativeness, temporal coherence, and plan accuracy). Our VG-TVP method outperforms unimodal baselines on the Daily-PP dataset.
\end{abstract}

%
 \begin{links}
      \link{Dataset}{https://github.com/mfurkanilaslan/VG-TVP}
 \end{links}

\begin{figure}[htb]
\begin{center}
    \includegraphics[width=\linewidth]{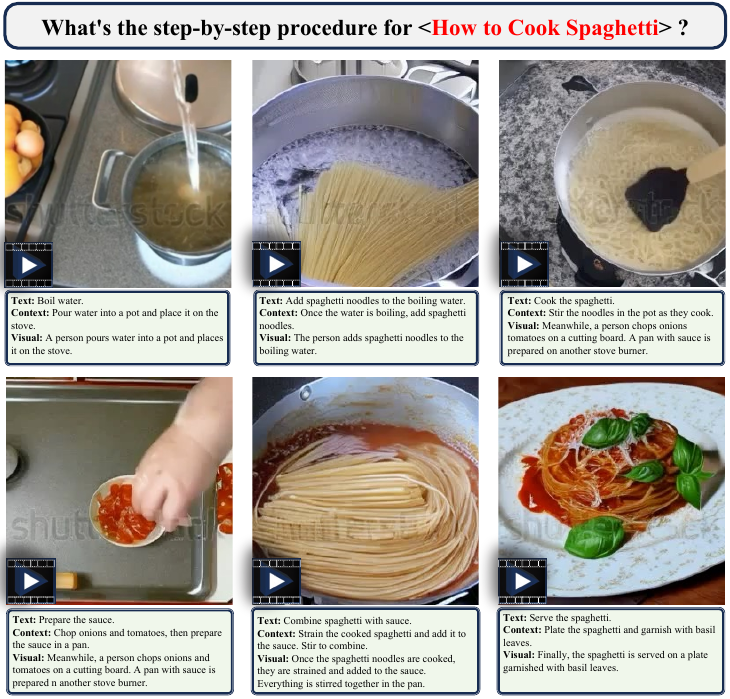}
\end{center}
  \caption{VG-TVP generates MPP with multiple steps for a high-level goal, supplying textual and visual guidelines.}
  \label{fig:teaser}
\end{figure}


\section{Introduction}
To acquire procedural knowledge, such as operating a machine, a person can refer to procedure plans, which specify the steps to achieve a task. Procedure plans may take different formats such as text, image, video, and a combination of them. Procedure Planning (PP) is the process of generating procedure plans. Depending on the modality of procedure plans, PP can be implemented using different methods and various sources of information. For example, LLMs have been used to generate procedure plans either in a zero-shot manner~\cite{DBLP:conf/icml/HuangAPM22} or by fusing information from various resources (e.g. WikiHow)~\cite{DBLP:journals/corr/abs-2305-01795}. 

Instructional videos (IVs) are a useful source of information on procedural knowledge. It provides a rich context of the task steps and effectively incorporates the temporal information essential for procedural knowledge learning. However, the quality of IVs might be inconsistent and it usually involves considerable effort of a human learner to understand the information and acquire the knowledge. To address this issue, one may resort to computational methods to parse the IVs and use the extracted visual elements to generate informative procedure plans. This necessitate several capabilities, such as semantics understanding ~\cite{DBLP:conf/cvpr/ZhukovACFLS19}, action-step prediction ~\cite{niu2024schema}, and scene understanding ~\cite{DBLP:conf/cvpr/ZhouMKSN23}. However, existing IVs may lack complete procedural information or contain visual content that does not align with the semantic plans. Hence, generating video content in association with the textual plans is desirable to form the final procedure plans.

In this paper, we aim to improve PP by generating multimodal content from different resources, including IVs. We are interested in enhancing human understanding by integrating visually grounded text and action-based video generations (Figure~\ref{fig:teaser}). Inspired by Text-to-Image Prompting (TIP)~\cite{DBLP:journals/corr/abs-2305-01795}, we cast the problem as MPP via Visually Grounded Text-Video Prompting (VG-TVP) (Figure~\ref{fig:Model_idea1}). VG-TVP generates video-enhanced action and state procedures given text descriptions of a task and IVs, which is in contrast to generating image plans and using their descriptions as text plans~\cite{DBLP:journals/corr/abs-2305-01795, DBLP:conf/cvpr/SoucekDWLS24}.

We anticipate that video-augmented procedure plans are advantageous to image-augmented ones~\cite{DBLP:journals/corr/abs-2305-01795, DBLP:conf/icml/RameshPGGVRCS21} because images focus on the static ``states'' of the task, whereas videos display dynamic changes of ``states'' with human-centred ``actions''. However, the challenges faced by TIP, including textual-visual informativeness, coherent temporal alignment, and high-level accurate plan generations~\cite{DBLP:journals/corr/abs-2305-01795}, become even more prominent in MPP. In particular, the \textit{framework} needs to ensure both temporal consistency (i.e. coherent temporal alignment between text and video plans) and spatial consistency (i.e. subsequent video steps must logically follow preceding ones in actions, objects, and contexts). Existing generative models, including LLMs and multimodal LLMs (MLLMs), cannot adequately address these challenges.

We propose a novel video-to-text$-$text-to-video (V2T-T2V) methodology that enhances the capability of LLMs for MPP tasks. Given the high-level goal-oriented task description, we first use LLMs to generate vanilla text plans. Meanwhile, we generate video captions from given IVs via V2T-Bridge (V2T-B). Subsequently, LLMs compile captions from various IVs to form a cohesive \textit{``Fusion of Captioning (FoC)''} (Figure~\ref{fig:FOC}) that adheres to the required steps. We leverage the \textit{LLMs'} capabilities by eliciting video captions and vanilla text plans to generate final revised text plans. Finally, video plans are generated by considering the generated text plans. Consequently, generated video plans are aligned with the generated text plans through VG-TVP.

Video-augmented PP is advantageous to IVs in several aspects. First, generated videos (a few seconds for each step) focus on the task steps, making them concise and relevant to the task. This may help alleviate the cognitive efforts of human learners than if they watch lengthy videos ($\approx5-10$ minutes) with much auxiliary information and time management. Second, the structured PP format could also improve the clarity and reliability of the MPP content, which facilitates self-paced learning. Third, using well-crafted prompts, we guide the video generation module to generate human-centred content, i.e. showing human actions whenever possible (Figure~\ref{fig:Qualitative}). This potentially helps reduce users' cognitive load owing to a smaller cognitive gap than if only state changes of the scene (without humans) are visualized.

VG-TVP addresses 3 major technical challenges: \textit{(1) Costly structure of foundational models (FM).} Although FMs are proficient in text generation, they must be trained with visual information. Training of data-hungry FMs from scratch demands high costs. In contrast, VG-TVP employs $3$ distinct components that strategically leverage each other's inputs and outputs, thereby eliminating the need to train separate models for specific tasks. 
\textit{(2) Alignment between procedures in IVs and generated text plans.} 
This challenge is addressed by FoC collaborating with VG-TVP alignment prompt. \textit{(3) Inadequacy of video captioning models to capture detailed MPP.} Captioning algorithms aim to capture the descriptions of the scenes, instead of detailed MPP. For instance, VLog~\cite{VLog_software} focuses on different aspects such as image, region, and the audio in the videos. Although it uses the latest LLMs, image, and audio models, it is not capable of combining the multimodal procedures for MPP tasks. 
A multimodal pre-trained video captioning model is proposed, focusing on dense events in the videos to capture descriptions in the same sequence {~\cite{DBLP:conf/cvpr/YangNSMPLSS23}. However, it only captures instructions, not coherent MPPs. In short, although existing methods are capable of captioning, additional information is required to generate MPPs.

The contributions of the study are four-fold. (1) We design a multi-modal framework, VG-TVP, that employs zero-shot prompting and avoids the costly training procedure. (2) We propose a novel method, FoC, which removes irrelevant contents in IVs, aligns mismatched steps between video captions and generates relevant text plans. They provide detailed MPP in the absence of IVs by exploiting video captions. (3) We propose Video-to-Text (V2T) and Text-to-Video (T2V) bridges to generate visually grounded text and video plans that are temporally coherent and accurate in planning. (4) We introduce a well-curated dataset called the Daily-Life Task Procedural Plans (Daily-PP) to mitigate the challenges in existing datasets for MPP content. VG-TVP's results show superior performance in the zero-shot setting compared to several baselines under the Daily-PP dataset.

\section{Related Works} 

\begin{figure*}
\begin{center}
\includegraphics[width=0.95\linewidth]{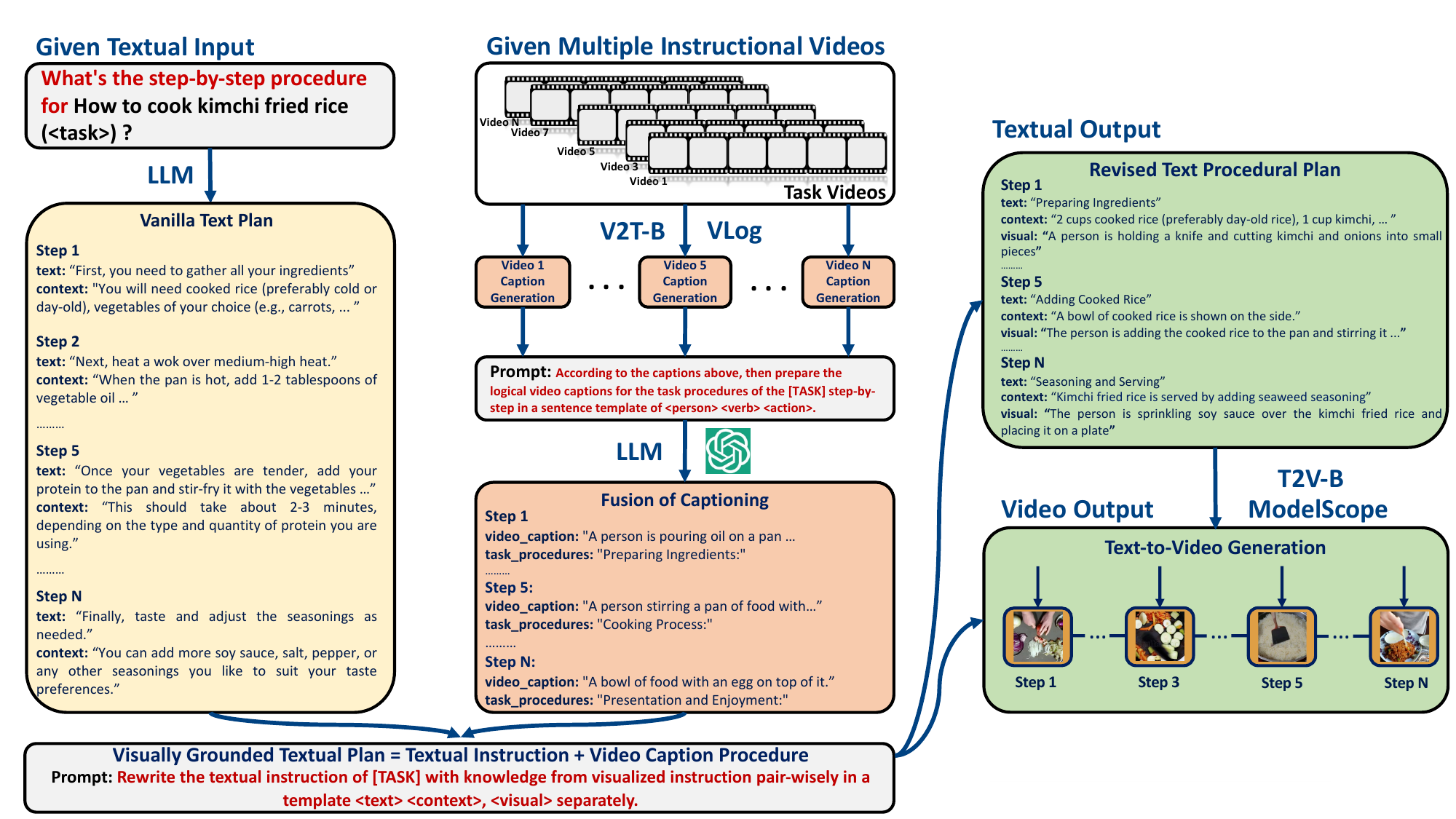}
\end{center}
   \caption{VG-TVP Model: Given the textual input and multiple instructional videos, VG-TVP generates visually grounded textual plans and video plans by using V2T-B and T2V-B. ChatGPT 3.5 is used to reorganize all captions to generate FoC.}
\label{fig:Model_idea1}
\end{figure*}

\subsection{Video Understanding for Procedural Planning} 
There has been remarkable progress achieved in understanding IVs via various problem statements~\cite{DBLP:conf/cvpr/KuehneAS14, DBLP:conf/iccv/MiechZATLS19, ilaslan-etal-2023-gazevqa, DBLP:conf/aaai/ZhouXC18, DBLP:conf/eccv/ChangHXAFN20}. While prior studies handle the problems via fully~\cite{DBLP:conf/iccv/ElhamifarN19, DBLP:conf/cvpr/ZhuY20a} or weakly-supervised~\cite{DBLP:conf/cvpr/ZhukovACFLS19, DBLP:conf/cvpr/0004HDDWJ22} or unsupervised settings~\cite{DBLP:conf/cvpr/AlayracBASLL16}, others focus on step discovery~\cite{DBLP:conf/cvpr/DvornikHZDWJ23}, prompting~\cite{DBLP:conf/iclr/LuFZXWEW23} or diffusion models~\cite{DBLP:journals/corr/abs-2309-07409, DBLP:conf/cvpr/WangW0023, DBLP:journals/corr/abs-2309-17444}. Unlike existing studies, we focus on generating visually grounded text-video pairs to obtain coherent MPPs.

\subsection{Multimodal Generation} 
Zero-shot planners~\cite{DBLP:conf/icml/HuangAPM22, DBLP:conf/iccv/SongSWCW023} or reasoners~\cite{DBLP:conf/nips/KojimaGRMI22} have leveraged the natural language generation capabilities of LLMs. While these efforts have yielded promising results, developing LLMs with multimodal content generation capabilities may lead to broader practical impacts. Although the performance of T2V models lagged behind that of Text-to-Image (T2I) models~\cite{DBLP:conf/icml/RameshPGGVRCS21}, recent advances have enhanced T2V performances~\cite{DBLP:journals/corr/abs-2211-11018, DBLP:conf/iclr/SingerPH00ZHYAG23, DBLP:conf/iccv/KhachatryanMTHW23, DBLP:conf/iccv/WuGWLGSHSQS23}. Existing models lack robustness as exhibited by LLMs in text generation, highlighting the need for advanced LLM-based frameworks for MPP tasks. Unlike existing models, VG-TVP leverages LLM capabilities for MPP tasks, offering dynamic, and comprehensive guidance through video generation in a human-centered manner.


\subsection{Integration of Visual Knowledge} IVs contain complex visual procedural knowledge that should be integrated into LLMs by reducing the complexity. Thus, it has been integrated with different methods such as usage of given images or generated texts as additional features~\cite{DBLP:conf/emnlp/YangYZWYC22, DBLP:journals/corr/abs-2305-01795}. However, image data results in inadequate descriptions of task actions since it provides only static information. To address that, VG-TVP harnesses IVs' captions to integrate visual knowledge into LLMs, leveraging their zero-shot reasoning capability. Moreover, in the PP tasks, integrated visual knowledge must be aligned for coherent outcomes~\cite{DBLP:conf/cvpr/ShenWE21, DBLP:conf/cvpr/DvornikHZDWJ23} which is a challenge for MPP tasks. To address that, we propose a method to combine video captions and vanilla text systematically.



\section{VG-TVP Approach}

We propose two key ideas in a methodology. The first idea involves generating MPPs for the tasks by exploiting their IVs, which is called ``SEEN'' (Idea 1). For the ``SEEN'' tasks, the model utilizes the relevant task IVs and captures their captions. Finally, video captions of these ``SEEN'' tasks and vanilla text plans are aligned to generate visually grounded text and video plans. The ``UNSEEN'' tasks (Idea 2) which the model has not previously encountered, involve generating MPPs for tasks that lack IVs. For example, ``\textit{Cooking Kimchi Fried Rice}'' and ``\textit{Cooking Szechuan Chicken}'' are two ``SEEN'' tasks under the Daily-PP dataset. We propose an exploration of ``\textit{Cooking Chicken Fried Rice}'' as an ``UNSEEN'' task by utilizing the captions of ``\textit{Cooking Kimchi Fried Rice}'' and \textit{"Cooking Szechuan Chicken}''.

\subsection{Problem Statement}
We formulate the problem as a visually grounded textual-video pairs alignment and generation task. $G^V$ is given multiple IVs which represent high-level goal videos, and $G^T$ is given textual input which denotes high-level goal texts, provided by the user in natural language. The model's output is a comprehensive multimodal procedural plan, represented as Goal Plan, $G^P$. It comprises the final sequence of visually-grounded text plan $TP = \{\{pt_1, pc_1\}, \{pt_2, pc_2\}, ... , \{pt_n, pc_n\}\}$  and video plan $VP = \{v_1, v_2, ... , v_n\}$ pairs. In this notation, $pt_i$ represents the text of the final revised textual plan and $pc_i$ denotes its corresponding context. Consequently, $G^P = \{TP, VP\}$ is generated by merging $3$ multimodal features: vanilla text generation, video captioning via V2T-Bridge, and video generation via T2V-Bridge.

\subsection{Method}

We leverage the zero-shot reasoning ability of LLMs to generate vanilla text plans by proposing a step-by-step prompting template. To enhance the MPP, we propose the VG-TVP model (Figure~\ref{fig:Model_idea1}), applying V2T-B and T2V-B for generating a comprehensive multimodal procedural plan.

\begin{table*}[htb]
\begin{center}
\begin{tabular}{ccccccccccccc}
\hline
\multicolumn{1}{l}{VG-TVP (Ours)} & \multicolumn{3}{c}{Textual Informative} & \multicolumn{3}{c}{Visual Informative} & \multicolumn{3}{c}{Temporal Coherence}  & \multicolumn{3}{c}{Plan Accuracy} \\
\multicolumn{1}{c}{versus}              & Win       & Tie         & Lose                 & Win       & Tie         & Lose            & Win       & Tie         & Lose          & Win       & Tie         & Lose            \\
\hline
Llama2-7B-q4            & \textbf{44.00}   & 40.00           & 16.00  & \textbf{74.00}     & 2.00      & 24.00     & \textbf{68.00} & 8.00           & 24.00 & \textbf{74.00} & 6.00  & 20.00           \\
Llama2-7B-q8            & \textbf{52.00}   & 22.00           & 26.00  & \textbf{62.00}     & 18.00     & 20.00     & \textbf{62.00} & 24.00          & 14.00 & \textbf{70.00} & 12.00 & 18.00           \\
Llama2-13B-q4           & \textbf{40.00}   & 24.00           & 36.00  & \textbf{46.00}     & 10.00     & 44.00     & \textbf{42.00} & 34.00          & 24.00 & 40.00          & 18.00 & \textbf{42.00}  \\
Llama2-13B-q8           & 20.00            & \textbf{48.00}  & 32.00  & \textbf{56.00}     & 18.00     & 26.00     & 36.00          & \textbf{44.00} & 20.00 & \textbf{46.00} & 22.00 & 32.00           \\
Mistral-7B-q4           & \textbf{58.00}   & 26.00           & 16.00  & \textbf{56.00}     & 12.00     & 32.00     & \textbf{52.00} & 20.00          & 28.00 & \textbf{48.00} & 28.00 & 24.00           \\
Mistral-7B-q8           & \textbf{54.00}   & 22.00           & 24.00  & \textbf{58.00}     & 14.00     & 28.00     & \textbf{46.00} & 34.00          & 20.00 & \textbf{56.00} & 16.00 & 28.00           \\
GPT3.5                  & \textbf{40.00}   & 24.00           & 36.00  & \textbf{56.00}     & 12.00     & 32.00     & \textbf{54.00} & 14.00          & 32.00 & \textbf{56.00} & 10.00 & 34.00           \\
TIP Model                  & \textbf{71.43}   & 21.43           & 7.14  & \textbf{57.14}     & 21.43     & 21.43     & \textbf{42.86} & 35.71          & 21.43 & \textbf{42.86} & 28.57 & 28.57           \\

\hline
\end{tabular}
\end{center}
\caption{Percentages of human evaluation comparisons between VG-TVP (Ours), baseline models by employing different LLMs (first 7 rows), and TIP~\cite{DBLP:journals/corr/abs-2305-01795} for SEEN tasks. Win represents the preferred option for VG-TVP.}
\label{table:1}
\end{table*}

\subsubsection{Vanilla Text Plan Generation.} 
The model generates vanilla text plans for the steps called Vanilla Textual Plan (VTP), VTP = $\{s_1, s_2, ..., s_n\}$ via vanilla prompt template function, $f_{prompt}(vanilla)$. This is a natural language format \textit{``What's the step-by-step procedure for $<$[TASK]$>$?''} used to elicit information from LLMs shown in Figure~\ref{fig:Model_idea1}. Inspired by the TIP approach ~\cite{DBLP:journals/corr/abs-2305-01795}, we adopt the zero-shot chain-of-thought approach~\cite{DBLP:conf/nips/KojimaGRMI22} for VTP generation. For example, for \textit{``Cooking Spaghetti''} task, the input text will be \textit{``What's the step-by-step procedure for How to Cook Spaghetti?''}. Each step represents a VTP paired with text and context at timestamp $i$, $s_i = \{t_i, ct_i\}$. While $T = \{t_1, t_2, ... , t_n\}$ represents the VTP-text, $CT = \{ct_1, ct_2, ..., ct_n\}$ denotes VTP-context generation. The initial text plan VTP is derived using $f_{prompt}(vanilla)$ focused on the specified $<$[TASK]$>$.

\begin{figure}[htb]
\begin{center}
  \includegraphics[width=0.78\linewidth, height=0.73\textwidth]{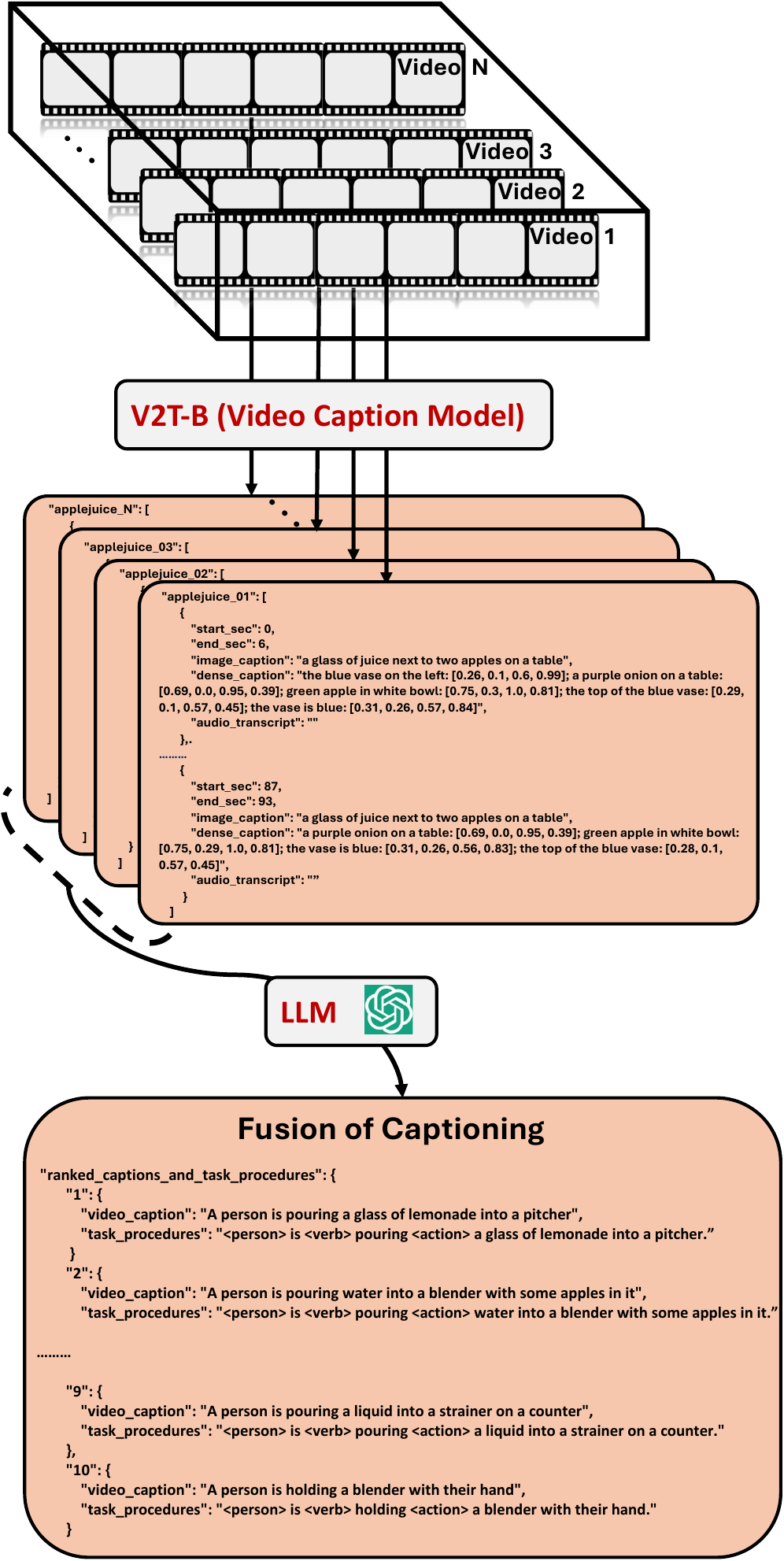}
  \caption{FoC captures and fuses IVs' captions. Then, it injects into the system by aligning them with vanilla textual.}
  \label{fig:FOC}
\end{center}
\end{figure}

\subsubsection{Fusion of Captioning.} 
We employ VLog model~\cite{VLog_software} which integrates the capabilities of ChatGPT~\cite{DBLP:journals/corr/abs-2303-08774}, BLIP2~\cite{DBLP:conf/icml/0008LSH23}, GRIT~\cite{DBLP:conf/eacl/DuRC21}, Whisper~\cite{DBLP:conf/icml/RadfordKXBMS23} models. VLog generates video captions from $3$ \textit{modalities}: image, region, and audio. Our framework excludes audio captions as some videos lacked speech or significant auditory content. Thus, we follow the visual information. V2T-B is the method that textualizes the scenes of IVs by using a video captioning algorithm. For example, in the ``apple juice'' task, given that the $3rd$ caption of the $1st$ IV represents the step ``wooden cutting board with sliced apples on it'' between $17-26$ seconds (sec), similar steps appear in the $4th$ IV at $39-50$ sec, and the $7th$ IV at $9-19$ sec in the $2nd$ step (not in $3rd$). In other IVs' captions, steps similar to ``pouring water into a blender'' are included in different steps. Therefore, FoC reorders and aligns the steps for vanilla textual matching (Figure 15 in the Appendix). FoC is insufficient for completing the MPP alone. It provides additional visual information to VG-TVP for generating MPP tasks. Consequently, we formulate the V2T-B as $FoC$ = $f_{captions}$ $\oplus$ \{($G^V$), $f_{prompt}(description)$\}. The \textit{Description} prompt is ``\textit{According to the captions above, then prepare the logical video captions for the task procedures of the $<$[TASK]$>$ step-by-step in a sentence template of $<$person$>$ $<$verb$>$ $<$action$>$?''}. Detailed video captions and FoC can be found in the supplementary materials.


\subsubsection{Visually-Grounded Textual and Video Plan Generation.} 
There has been a shift toward leveraging LLMs for visual generations~\cite{DBLP:journals/corr/abs-2310-05737, DBLP:conf/cvpr/BlattmannRLD0FK23}, moving beyond the conventional use of diffusion models. Specifically,~\cite{DBLP:journals/corr/abs-2309-15091} equipped with image annotations and a GPT4-based video planner, is prone to generating unnecessary instructions for PP tasks. For instance, \textit{``A hungry cat is finding food.''} prompt in~\cite{DBLP:journals/corr/abs-2309-15091} leads LLMs to segment the generated narrations into scenes. VG-TVP focuses on generating MPP to assist humans rather than storytelling or directing videos. VG-TVP's $Goal Plan$ is the final sequence of visually-grounded text plan (TP) and video plan (VP) pairs, $G^P = \{TP, VP\}$. However, multimodal generation brings an alignment challenge. Thus, we propose FoC collaborating with VG-TVP alignment prompt to revise VTPs and tackle this challenge.

Text-video alignment is essential for maintaining consistency and coherence between textual and visual plans. It synchronizes procedural steps across modalities, enabling users to correlate textual instructions with relevant generated videos and follow the instructions accurately. This aids comprehension and enhances the learning experience by offering unified and coherent instructions. We formulate the prompt alignment with the integration of VTPs and FoC via a prompt template: $TP$ = $f_{prompt}(alignment)$  $\oplus$ $\{(FoC), (VTP)\}$. \textit{Alignment} prompt represents ``\textit{Rewrite the step-by-step procedures of $<$[TASK]$>$ by using video captions pair-wisely in a template $<$text$>$, $<$context$>$ and $<$visual$>$ separately.''}. We follow this alignment to generate the relevant task videos which compose the Video Plan, $VP$ = (T2V-B) $\oplus$ ($TP$). $VP$ is derived by employing T2V-B which exploits a sequence of visually-grounded text plan $TP$. Then, we generate relevant tasks' video plans by using ModelScope~\cite{DBLP:journals/corr/abs-2308-06571}. Consequently, a visually grounded approach leverages both text and visual information to tackle the alignment and coherency challenges.

\begin{table*}[htb]
\begin{center}
\begin{tabular}{ccccccccccccc}
\hline
\multicolumn{1}{l}{VG-TVP (Ours)} & \multicolumn{3}{c}{Textual Informative} & \multicolumn{3}{c}{Visual Informative} & \multicolumn{3}{c}{Temporal Coherence}  & \multicolumn{3}{c}{Plan Accuracy} \\
\multicolumn{1}{c}{versus}           & Win       & Tie         & Lose                 & Win       & Tie         & Lose            & Win       & Tie         & Lose          & Win       & Tie         & Lose            \\
\hline
Llama2-7B-q4    & \textbf{40.00}   & 26.67           & 33.33  & \textbf{60.00}     & 26.67     & 13.33     & \textbf{60.00}    & 26.67            & 13.33             & \textbf{66.67}  & 20.00   & 13.33             \\
Llama2-7B-q8    & \textbf{66.67}   & 6.67            & 26.67  & \textbf{53.33}     & 26.67     & 20.00     & \textbf{53.33}    & 26.67            & 20.00             & \textbf{60.00}  & 6.67    & 33.33             \\
Llama2-13B-q4   & 26.67            & \textbf{53.33}  & 20.00  & \textbf{46.67}     & 13.33     & 40.00     & \textbf{53.33}    & 26.67            & 20.00             & \textbf{40.00}  & 33.33   & 26.67             \\
Llama2-13B-q8   & \textbf{53.33}   & 40.00           & 6.67   & \textbf{60.00}     & 33.33     & 6.67      & \textbf{80.00}    & 0.00             & 20.00             & \textbf{66.67}  & 13.33   & 20.00             \\
Mistral-7B-q4   & \textbf{40.00}   & \textbf{40.00}  & 20.00  & \textbf{53.33}     & 26.67     & 20.00     & \textbf{46.67}    & 46.67            & 6.67              & \textbf{40.00}  & 53.33   & 6.67              \\
Mistral-7B-q8   & \textbf{73.33}   & 20.00           & 6.67   & \textbf{86.67}     & 0.00      & 13.33     & \textbf{66.67}    & 26.67            & 6.67              & \textbf{73.33}  & 20.00   & 6.67              \\
GPT3.5          & 13.33            & \textbf{80.00}  & 6.67   & \textbf{60.00}     & 6.67      & 33.33     & 20.00             & \textbf{40.00}   & \textbf{40.00}    & 26.67           & 33.33   & \textbf{40.00}    \\
\hline
\end{tabular}
\end{center}
\caption{Percentages of human evaluation comparisons between VG-TVP (Ours) and baseline models by employing different LLMs (q: quantization) for UNSEEN tasks. WIN, and LOSE represent the better and worse results of VG-TVP, respectively.}
\label{table:2}
\end{table*}

\section{Experiments}
Baselines utilize text/context to generate text-aligned inclusive vanilla video plans. On the other hand, VG-TVP exploits vanilla text and FoC to generate visually grounded MPP content to assist individuals. We use a Win-Tie-Lose comparison on 50 seen and 15 unseen tasks, involving 28 human subjects for benchmarking. VG-TVP generated 2,504 videos for ``seen'' and 687 for ``unseen'' tasks, while baselines produced 2,701 and 681 vanilla textual videos, respectively. We included $2$ tasks from WikiPlan\&RecipePlan, TIP model~\cite{DBLP:journals/corr/abs-2305-01795}, to facilitate a fair benchmarking. Moreover, we conduct one more comparison with $2$ different prompts (in the Appendix). This comparison aims to compare the effectiveness of injecting human orientation into text and video plans with VG-TVP against prompting. The qualitative results display that human orientation is more effectively integrated with VG-TVP. Additionally, we measure the textual relevance between generated baselines' and VG-TVP's text plans with reference text plans using BLEU~\cite{10.3115/1073083.1073135}, and METEOR~\cite{banerjee-lavie-2005-meteor}. Finally, we design an LLM (via ChatGPT4o) (with Socratic Method ~\cite{chang2023promptinglargelanguagemodels}) evaluation protocol to evaluate baselines and VG-TVP on $4$ aspects as in the human evaluation metric. Each scored out of $25$ points, to assess the quality of task plans. The details are in the Appendix.

\subsection{Daily-PP Dataset}

Existing datasets are not suitable for the MPP content generation. Specifically, CrossTask~\cite{DBLP:conf/cvpr/ZhukovACFLS19}, lacked specified task patterns, making it difficult to use for PP analysis. COIN~\cite{DBLP:conf/cvpr/TangDRZZZL019}, does not align with the structure needed for our research. Similarly, ProceL~\cite{DBLP:conf/iccv/ElhamifarN19} required updates to meet the specific demands of IV analysis. WikiPlan\&RecipePlan~\cite{DBLP:journals/corr/abs-2305-01795} is not precisely tailored for following a structured task sequence. Thus, we curated a new dataset - called \textit{Daily-Life Task Procedural Plans (Daily-PP)} - to better align with MPP content, drawing inspiration from the strengths and addressing the limitations of existing collections.

The Daily-PP consists of 5 domains (Breakfast, Dinner, Drink, Hobby\&Crafts, and Home\&Garage), 50 seen tasks, and 15 unseen tasks. Seen tasks include 7 or 10 IVs from YouTube, depending on the video density for each task. Moreover, $3$ domains (Breakfast, Drink, and Dinner) have unseen tasks such as egg benedict, carrot mango lassi, and chicken fried rice. Unseen tasks are those without any IVs. The model generates their MPPs by using their vanilla text plan with $2$ relevant seen tasks' video captions. For example, VG-TVP uses the video captions of \textit{carrot juice} \& \textit{mango lassi} tasks to generate the MPP of \textit{carrot mango lassi}. The Daily-PP dataset structure is shown in the Appendix.


\subsection{Human Evaluation Metric}

Existing metrics such as BLEU and METEOR evaluate text similarity by comparing generated texts with reference texts. However, they have limitations in MPP tasks. In cases where tasks do not have strict laws or steps, it cannot be assumed there is a single ground truth (GT). Daily life tasks such as ``Hanging a Mirror'', and ``Cooking Pancakes'' lack definitive instructions and can vary widely. They cannot be deemed as having only one correct GT. Thus, a human evaluation survey is the optimal metric for assessing tasks aimed at generating MPPs. An example of a survey is in the Appendix. 


\subsection{Results}

\subsubsection{Baselines.} We employ 7 different LLMs to generate vanilla text plans ~\cite{DBLP:journals/corr/abs-2307-09288, DBLP:journals/corr/abs-2310-06825, DBLP:conf/nips/BrownMRSKDNSSAA20} which are (1) LLama2-7B-q4, (2) LLama2-7B-q8, (3) LLama2-13B-q4, (4) LLama2-13B-q8, (5) Mistral-7B-q4, (6) Mistral-7B-q8 and (7) ChatGPT3.5. 


\begin{table}
\begin{tabular}{ccccc}
\hline
Models & \multicolumn{2}{c}{BLEU} & \multicolumn{2}{c}{METEOR}  \\
              & Base.        & VG-TVP         & Base.         & VG-TVP           \\
\hline
Llama2-7B-q4  & 0.013           & 0.013          & \textbf{0.089}   & 0.082            \\
Llama2-7B-q8  & \textbf{0.014}  & 0.011          & \textbf{0.082}   & 0.075            \\
Llama2-13B-q4 & \textbf{0.012}  & 0.011          & 0.067            & \textbf{0.071}   \\
Llama2-13B-q8 & \textbf{0.014}  & 0.012          & \textbf{0.082}   & 0.073            \\
Mistral-7B-q4 & 0.012           & \textbf{0.014} & 0.091            & \textbf{0.093}   \\
Mistral-7B-q8 & 0.013           & \textbf{0.015} & 0.084            & \textbf{0.094}   \\
GPT3-5        & 0.017           & \textbf{0.024} & 0.072            & \textbf{0.085}   \\
\hline
\end{tabular}
\caption{Automatic evaluations on 50 seen tasks from Daily-PP. Generated Baselines' (Base.) and VG-TVP's text plans are compared with the reference textual.}
\label{table:3}
\end{table}

\subsubsection{Quantitative Analysis.}

The performance of VG-TVP in ``SEEN'' tasks is shown in Table~\ref{table:1}. There are only two ``TIE'' results in the comparison between VG-TVP and Llama2-13B-q8, specifically for textual informativeness (48.00\%) and temporal coherence (44.00\%). The highest preference rate of VG-TVP is observed as 58.00\% against Mistral-7B-q4. Benchmarking with the TIP model demonstrates that VG-TVP achieved better textual informativeness with a score of 71.43\%. In terms of visual informativeness, VG-TVP's performance ranges from a lower bound of 46.00\% to a higher bound of 74.00\%. VG-TVP, without losing any evaluation against the baseline models, achieves the highest average score in this challenge. In comparison with the TIP model, VG-TVP generates better visual plans with a 57.14\% preference score. In temporal coherence, VG-TVP obtains a higher preference rate of 36.00\% against Llama2-13B-q8, while achieving a preference parity of 44.00\% among subjects. In the benchmarking comparison with the TIP model, VG-TVP reaches a better rate with 42.86\%. In plan accuracy, VG-TVP’s ratio is only 2.00\% behind Llama2-13B-q4, yet it achieves successful ratios against other baselines. The benchmarking with the TIP model, VG-TVP achieves more accurate plans with 42.86\%. 

For the ``UNSEEN'' tasks, unlike the TIP model, VG-TVP can combine different task information to generate multimodal procedural plans for new, unseen content that lacks IVs or additional information. Therefore, the TIP model is not included in the detailed benchmarking, in Table~\ref{table:2}. In terms of textual informativeness, VG-TVP is not preferred over Llama2-13B-q4 and GPT3.5. However, the majority of subjects rated the comparison as ``TIE''. In visual informativeness, VG-TVP achieves superior preference rates, ranging from 46.67\% to 86.67\%, against all baselines. Regarding temporal coherence and plan accuracy, VG-TVP obtains better ratios than all models except GPT3.5. Consequently, VG-TVP's superior performance can be seen from higher preferences over baselines across various tasks and models, particularly in visual informativeness, showing its effectiveness in MPP.

Table~\ref{table:3} shows the textual relevance evaluations of baseline and VG-TVP's text plans using BLEU and METEOR on 50 seen tasks from the Daily-PP dataset. Except for the LLama2-13B-q4 on the METEOR and the LLama2-7B-q4 on the BLEU, baseline-generated text plans outperform via LLama2-7B-q8, LLama2-13B-q4, and LLama2-13B-q8 models. For Mistral-7B-q4, Mistral-7B-q8, and GPT3.5, VG-TVP text plans perform better in both metrics. These results cannot show the semantic performance of the text plans because a single GT for daily life tasks is not adequate to cover the MPP content. Therefore, human evaluation is the optimum metric for MPP tasks to assist people.

\begin{figure}
\begin{center}
    \includegraphics[width=\linewidth]{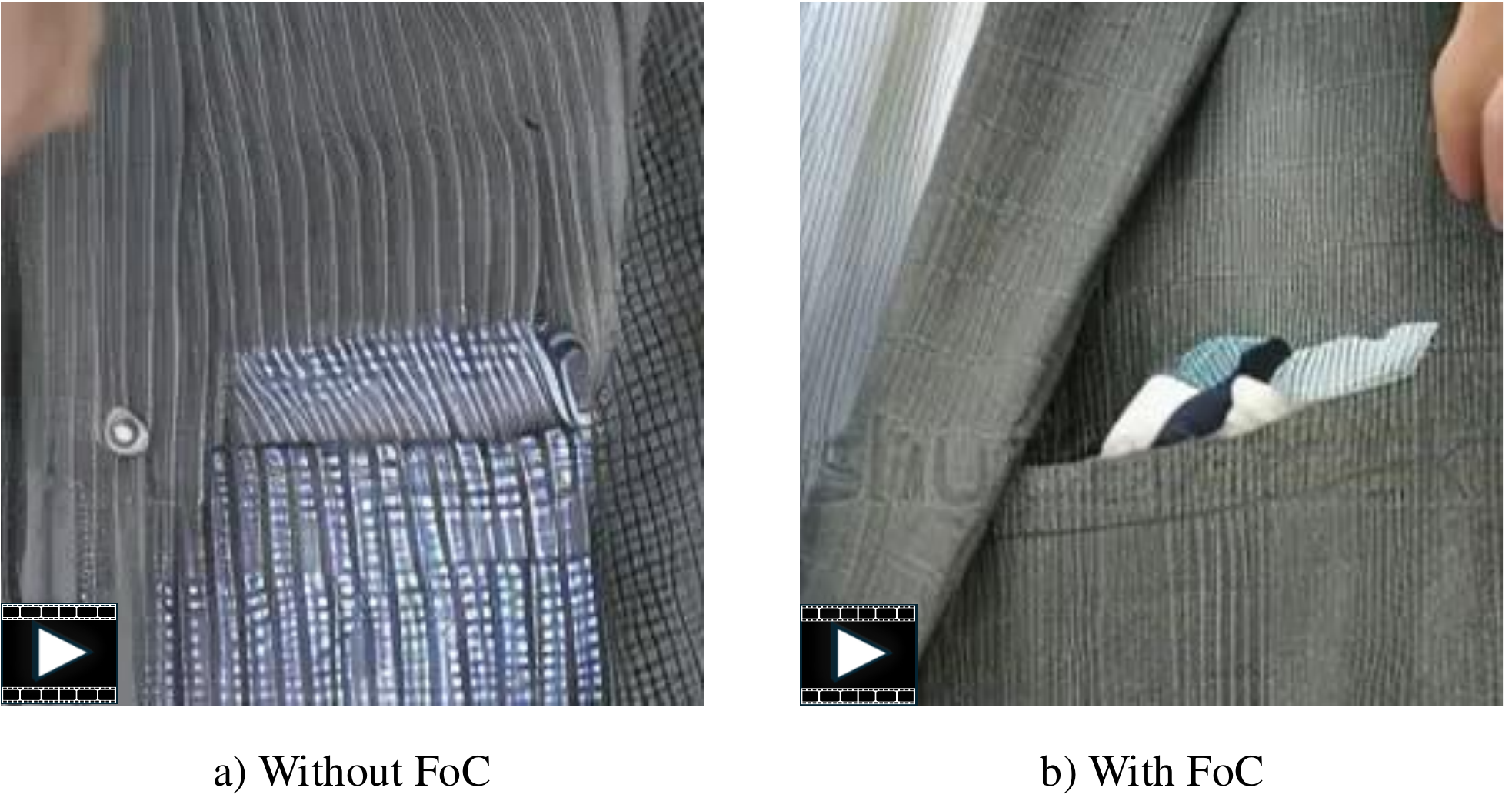}
\end{center}
  \caption{Impact of FoC on the task, ``How to Fold the Presidential Pocket Square?''.}
  \label{fig:FoCw2}
\end{figure}

\begin{figure*}[htb]
\begin{center}
\includegraphics[width=\linewidth]{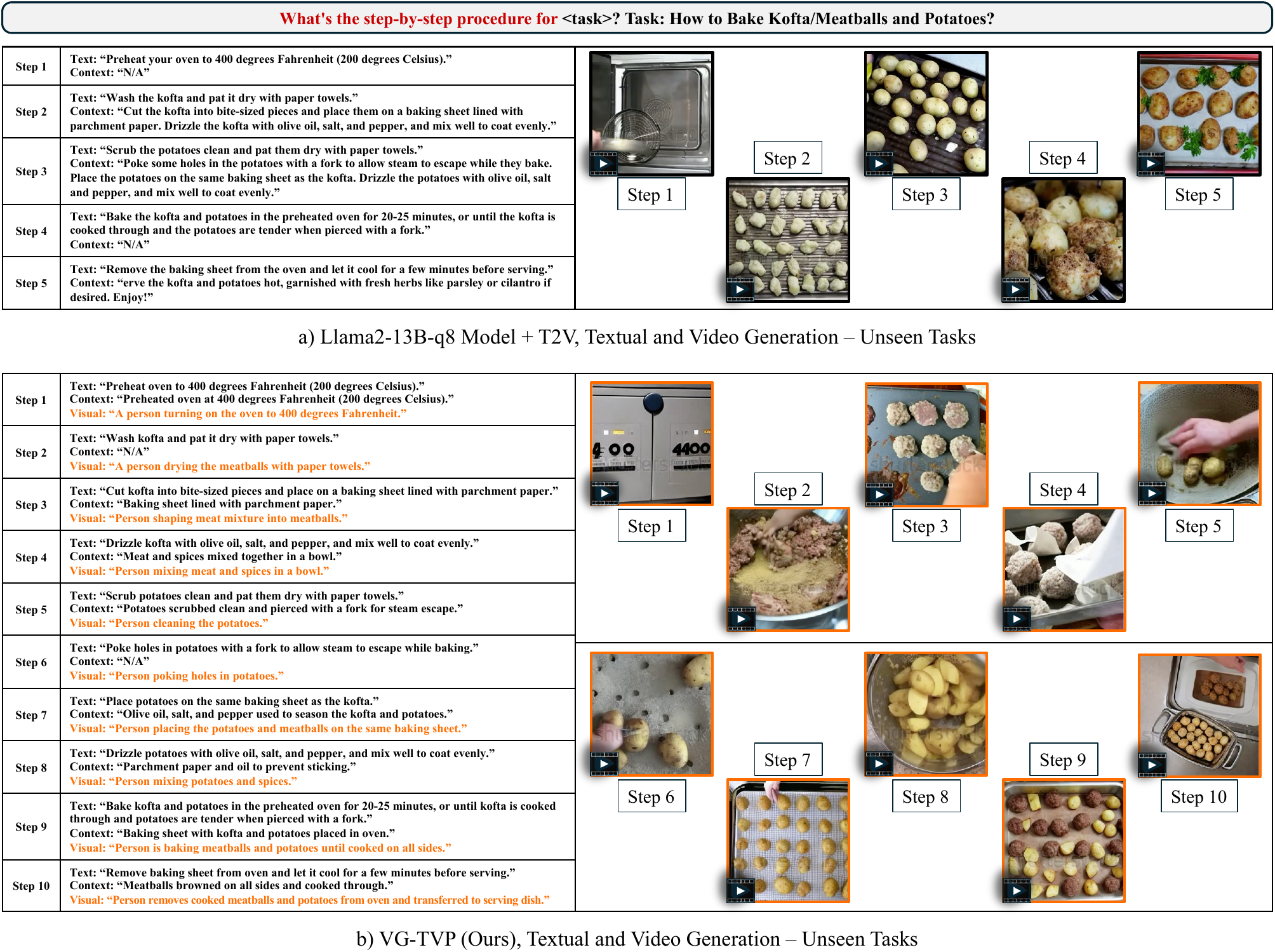}
\end{center}
    \caption{Qualitative comparison between Llama2-13B-q8 Model and VG-T2V (Ours). Visuals (orange) are used to generate video plans. VG-TVP can increase the number of steps to generate the MPP more informative and accurate.}
\label{fig:Qualitative}
\end{figure*}

\subsubsection{Qualitative Analysis.}

The generated vanilla text plans successfully verbalize procedural information across many tasks. However, using vanilla text plans may not achieve the same efficiency for visuals while generating videos. We hypothesize that augmenting these vanilla textual with IVs can enhance MPP, thereby more effectively assisting individuals. Our results and comparative analyses with VG-TVP and baselines confirm this hypothesis. We generate baseline models' videos using the text-context structure of vanilla textual. Figure~\ref{fig:Qualitative} provides a qualitative analysis through a visual example using the Llama2-13B-q8$+$T2V. The unseen task, ``How to bake kofta/meatballs and potatoes?'' combines ``Pan-Cooking Kofta/Meatballs'' and ``Cooking French Fries''. Figure~\ref{fig:Qualitative} a) displays the vanilla text and video plan generations, where the baseline model outlines five steps to complete the task. In Figure~\ref{fig:Qualitative} b), text\&video plans are generated using VG-TVP, and video plans are highlighted through visual prompts. VG-TVP enhances textual information, thereby improving the quality of generated video plans.

\subsubsection{Ablations.}
FoC is the key technical component of VG-TVP which captures and fuses IVs' captions to integrate into the VG-TVP. For example, Figure~\ref{fig:FoCw2} a) is generated based on \textit{Text: ``Finally, tuck the ends of the pocket square into your pocket to create a neat and tidy appearance''}, \textit{Context: ``Remember, the key to folding a pocket square is to be consistent and precise in your folds and to make sure the edges are aligned, and the corners are squared off.''}. The FOC leverages textual to generate a ``Visual'' prompt as \textit{``A person tucking the ends of a folded pocket square into their pocket, creating a neat and tidy appearance.''} that is used to generate the video shown in Figure~\ref{fig:FoCw2} b). Unlike baseline's generation (Figure~\ref{fig:FoCw2}a), VG-TVP generates a video with sharp lines of the suit jacket and folded square inside the pocket (Figure~\ref{fig:FoCw2}b). Consequently, FoC improves the impact of VG-TVP to generate more plan-accurate and informative visuals.


\section{Conclusion and Future Work}
This paper presents a novel approach to MPP through the development of LLM-powered frameworks, addressing the complexities of generating cohesive PPs for both seen and unseen tasks. By leveraging the capabilities of LLMs, VG-TVP enhances the coherence and consistency of PPs, demonstrating the efficacy of integrating visually grounded text and action-based video generations to enhance human assisting. Daily-PP dataset represents a significant stride towards overcoming the limitations of existing datasets, providing a more structured and comprehensive resource for evaluating MPP. VG-TVP may serve as an effective model for future research on leveraging multimodal information to enhance human learning experiences.

\section*{Acknowledgements}

This research is supported by the Singapore International Graduate Award (SINGA) Scholarship.

\bibliography{aaai25}

\appendix
\section{Appendix}
\label{sec:appendix}

VG-TVP focuses on generating multimodal text and video plans by capturing "state$+$action" from instructional videos (IVs). Despite having a limited number of IVs, VG-TVP generates textual and video plans for unseen tasks using relevant combinations. This capability indicates that a potential limitation has been addressed to a certain extent. Empirical results demonstrate VG-TVP's ability to generate informative, temporally coherent, and accurate MPP content. We propose a new framework for future research on leveraging multimodal information to enhance human learning experiences. It might serve as an effective and guiding resource for future studies, addressing key questions from educational psychology to cognitive load theory, such as \textit{"How much do the generated instructions boost the humans' success rate?"}, with the advancement of AI models. 

For fair comparisons with the baselines, inference parameters of all models have been set as 0.8 "temperature", 40 "top-k", 0.05 "min-p", and 0.095 "top-p" sampling, 512 for "prompt evaluation batch size (n-batch)", and 4096 tokens for "context length (n-ctx)". Moreover, the system prompt has been set as "You are a helpful AI assistant.".

%

\subsection{Daily-PP Dataset}

The Daily-PP dataset (Figure~\ref{fig:Dataset}) consists of $5$ domains which are Breakfast, Dinner, Drink, Hobby\&Crafts, and Home\&Garage. It includes 50 seen tasks and 15 unseen tasks, each with 7 to 10 IVs sourced from YouTube, reflecting the task’s commonality. Unlike unseen tasks, the seen tasks are directly supported by these IVs. Unseen tasks are conceptually modeled through captions from related seen tasks. For example, carrot juice and mango lassi IVs are used to formulate the MPP for carrot mango lassi.

\subsubsection{Statistical Analysis.} $700$ and $300$ unique action verbs were generated for seen and unseen tasks, respectively. Specifically, the verbs for seen tasks were utilized approximately 3,000 times in "text," over 8,000 times in "context," and more than 5,000 times in "visual" settings (in total $>$ $16,000$). For unseen tasks, the verbs were used over $700$ times in "text," more than $1,900$ times in "context," and over $1,400$ times in "visual" settings, (in total $>$ $4,000$).

For seen tasks, more than $3,000$ unique words were generated. These words were used over $19,000$ times in "text," more than $50,000$ times in "context," and over $27,000$ times in "visual" settings, totaling nearly $100,000$ instances. For unseen tasks, over $1,200$ unique words were generated. These words were used more than $5,200$ times in "text," over $13,000$ times in "context," and more than $7,000$ times in "visual" settings, (in total $>$ $26,000$).

VG-TVP generated over $700$ action verbs for both seen and unseen tasks. $3,000$ and $1,500$ unique words were generated for seen and unseen tasks, respectively. These words were used over $83,000$ times for seen tasks and more than $33,000$ times for unseen tasks. The detailed graphs to show the top 30 "verbs and words" frequencies and word cloud distributions for the seen and unseen tasks at the baselines, and VG-TVP are displayed below from Figures~\ref{fig:output_seen_top30words_in_text} to~\ref{fig:wdc_output}.

\begin{figure}
  \includegraphics[width=\linewidth]{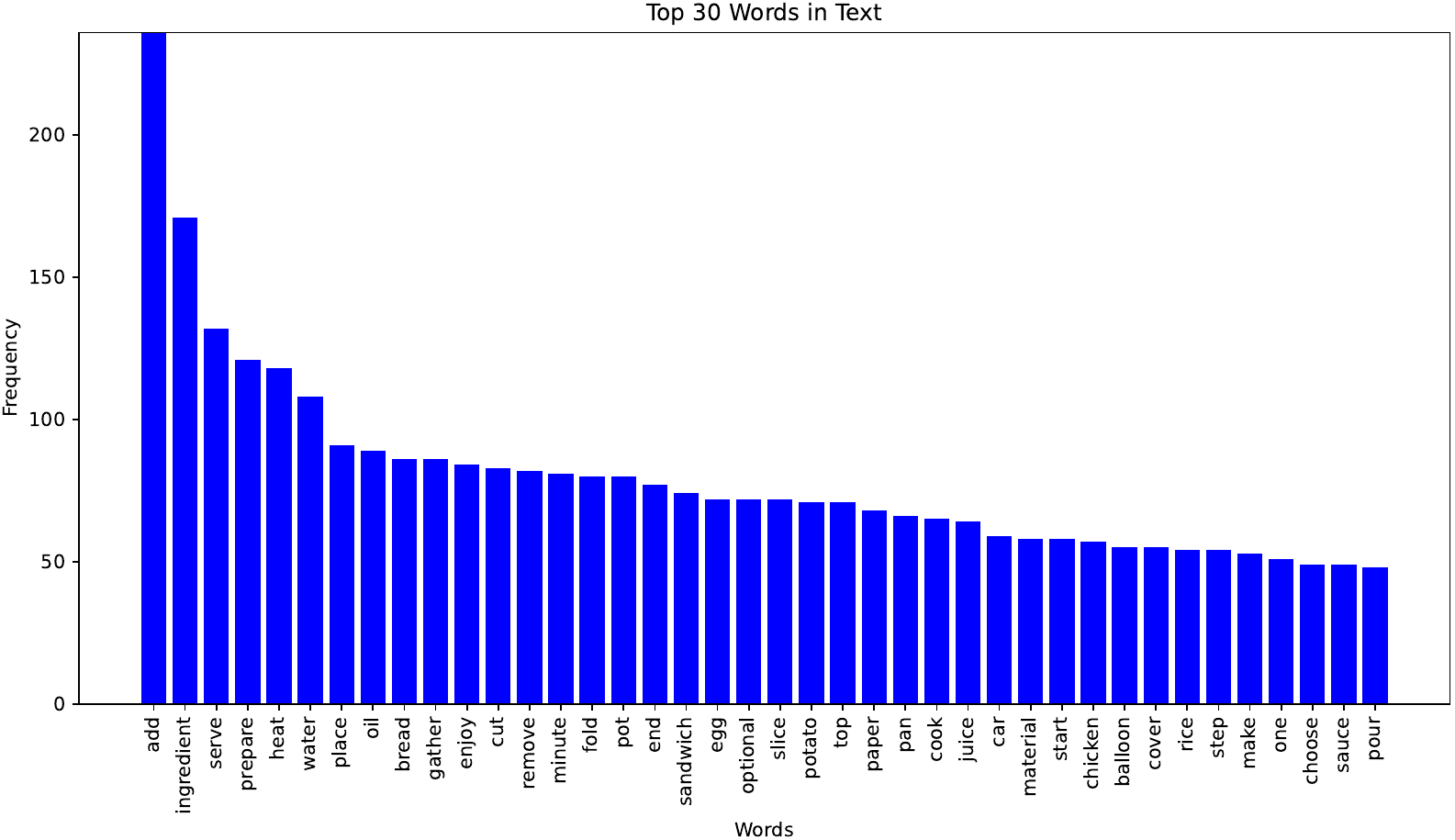}
  \caption{Top 30 Words in "Text", Seen Tasks, VG-TVP}
  \label{fig:output_seen_top30words_in_text}
\end{figure}

\begin{figure}
  \includegraphics[width=\linewidth]{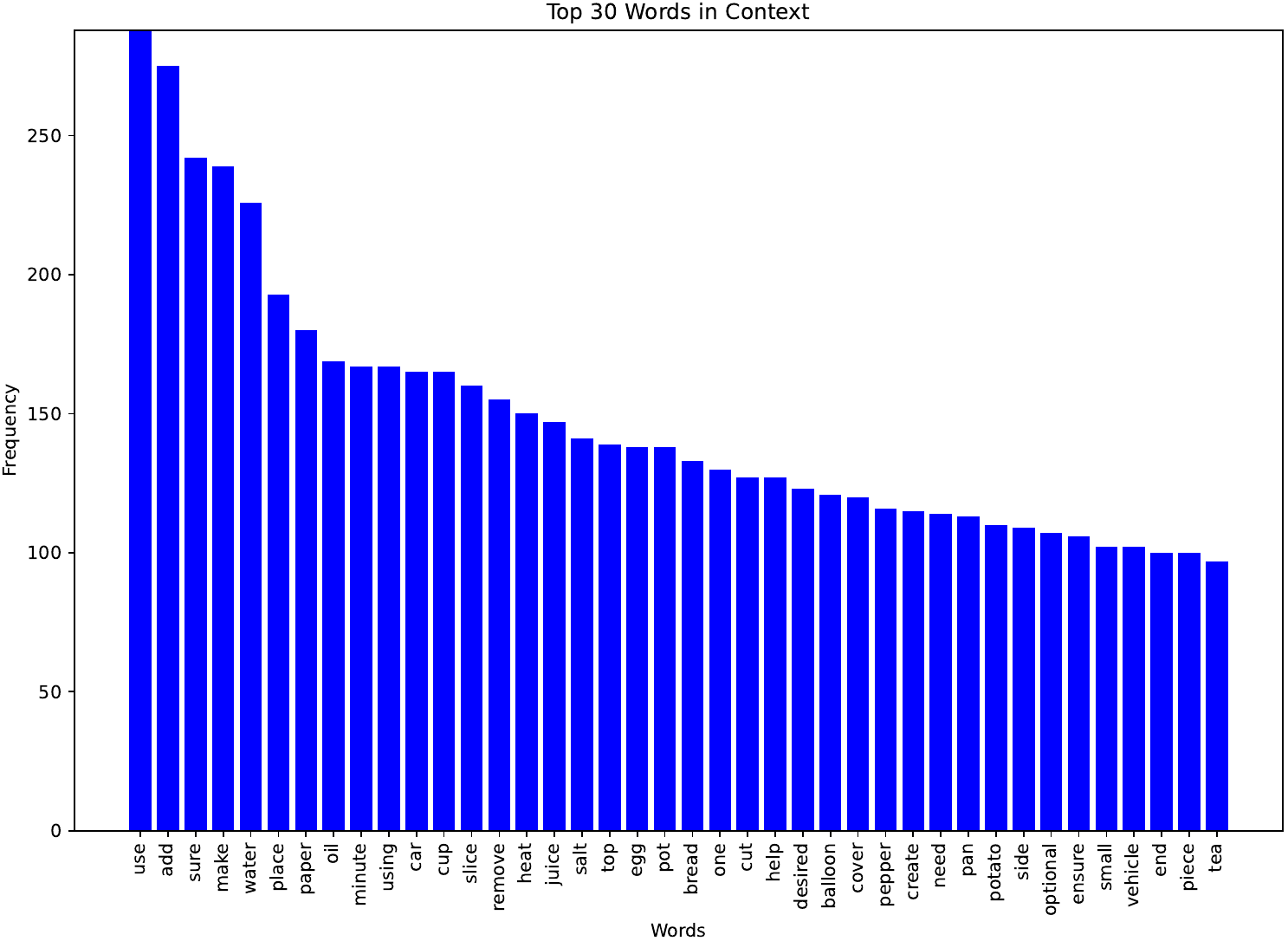}
  \caption{Top 30 Words in "Context", Seen Tasks, VG-TVP}
  \label{fig:output_seen_top30words_in_context}
\end{figure}

\begin{figure}
  \includegraphics[width=\linewidth]{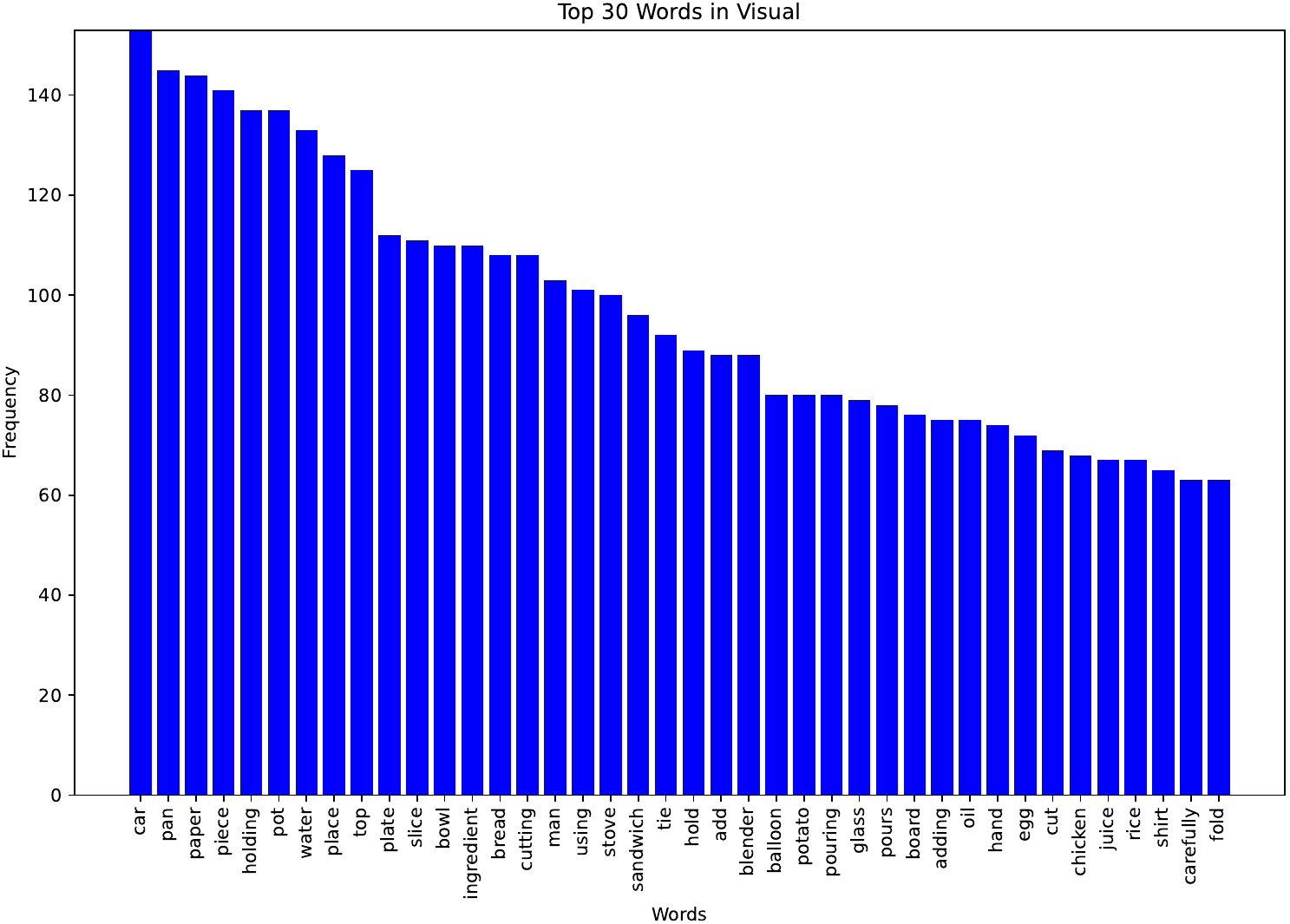}
  \caption{Top 30 Words in "Visual", Seen Tasks, VG-TVP}
  \label{fig:output_seen_top30words_in_visual}
\end{figure}

\begin{figure}
  \includegraphics[width=\linewidth]{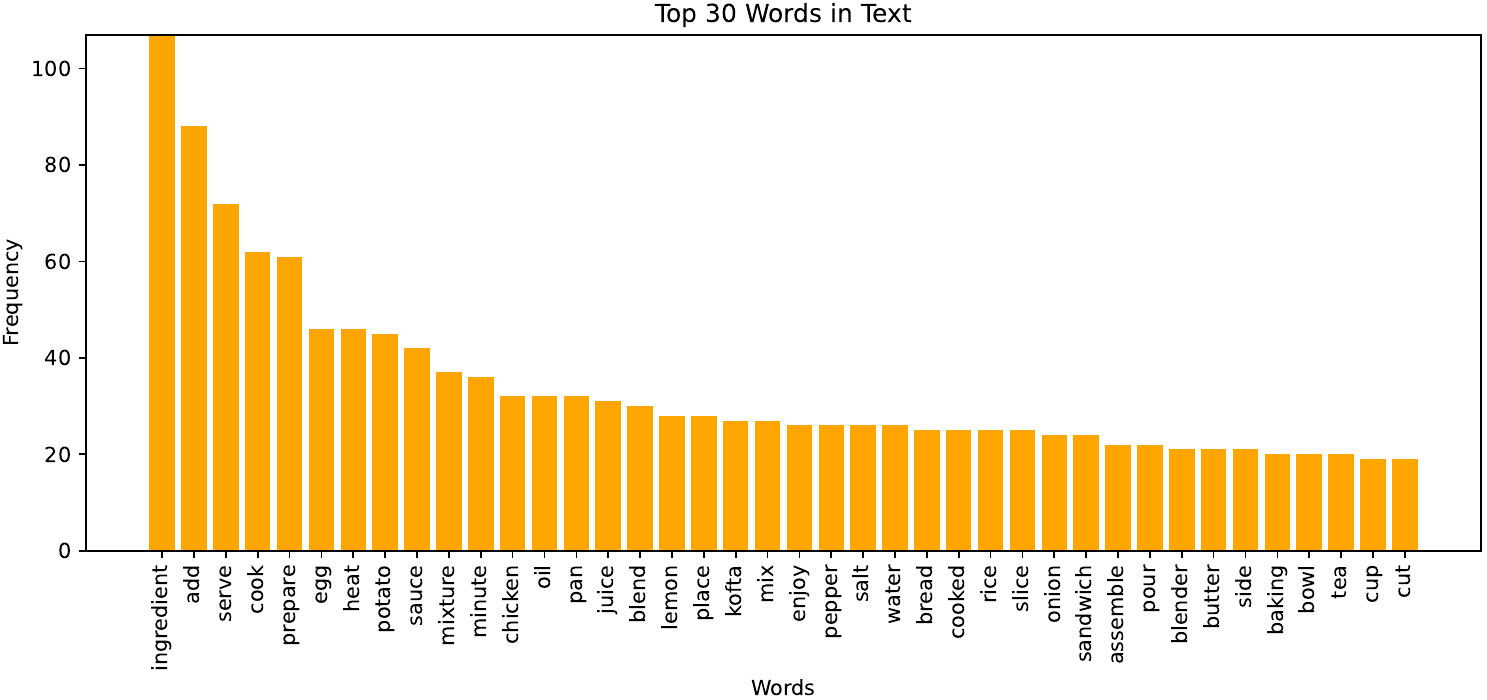}
  \caption{Top 30 Words in "Text", Unseen Tasks, VG-TVP}
  \label{fig:output_unseen_top30words_in_text}
\end{figure}

\begin{figure}
  \includegraphics[width=\linewidth]{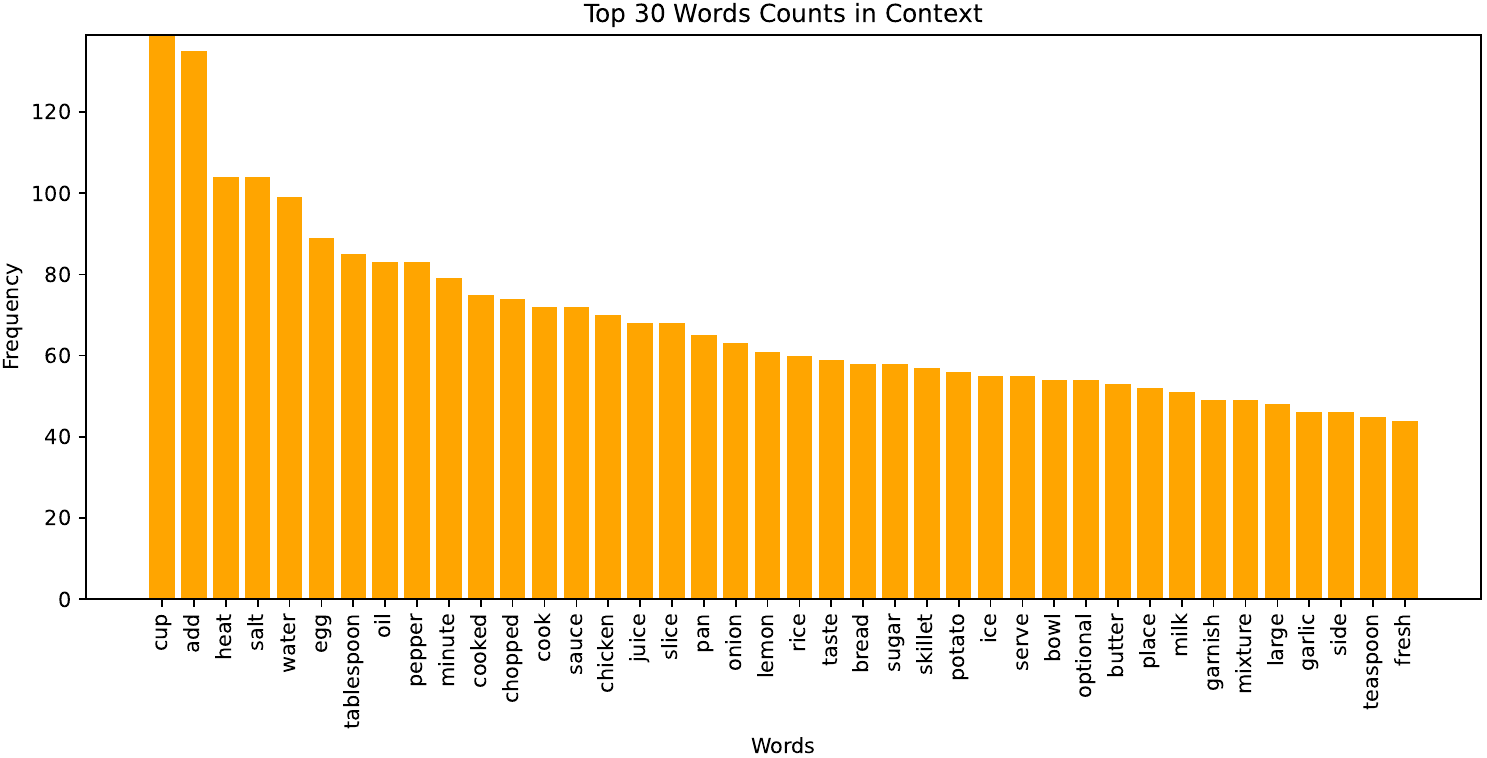}
  \caption{Top 30 Words in "Context", Unseen, VG-TVP}
  \label{fig:output_unseen_top30words_in_context}
\end{figure}

\begin{figure}
  \includegraphics[width=\linewidth]{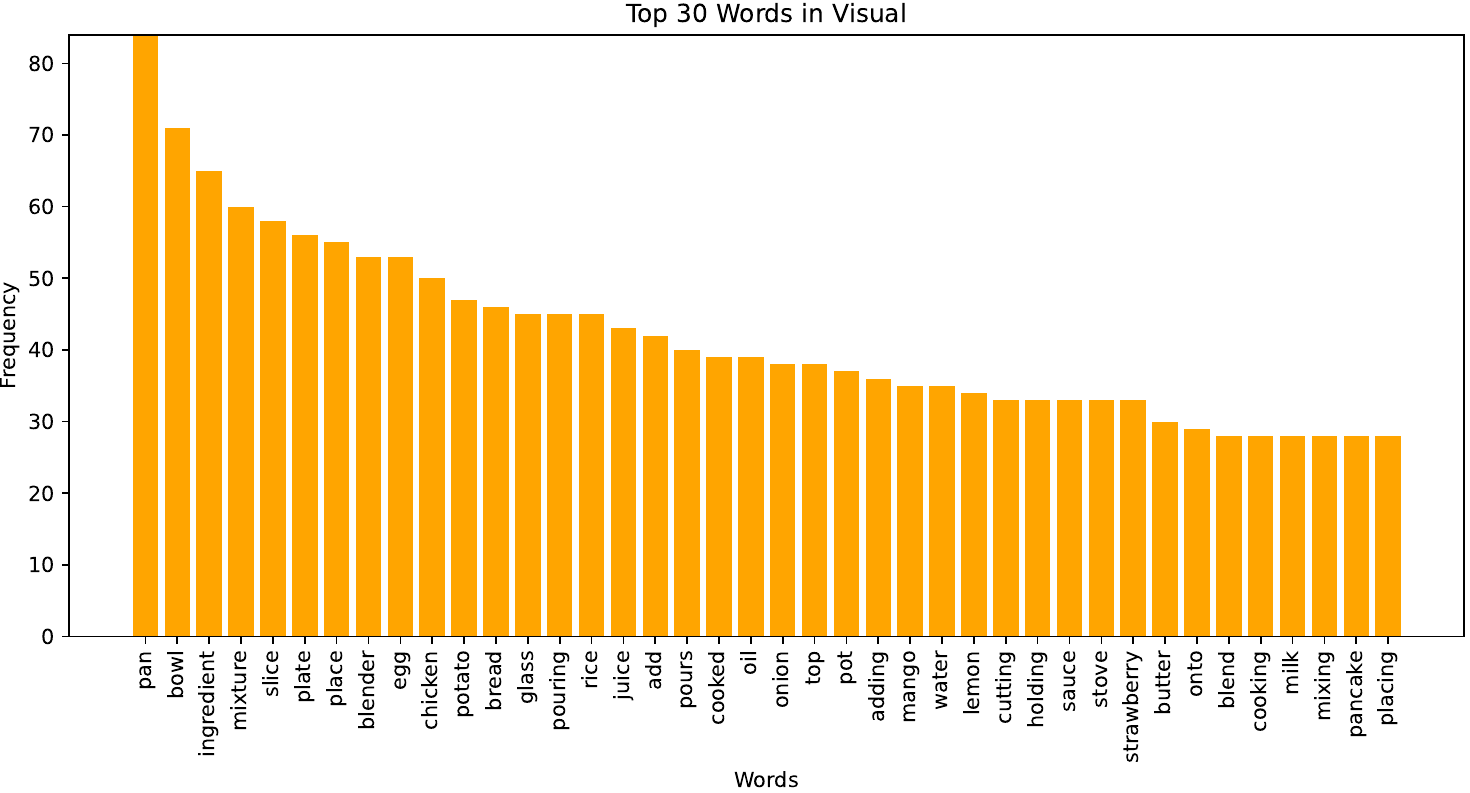}
  \caption{Top 30 Words in "Visual", Unseen Tasks, VG-TVP}
  \label{fig:output_unseen_top30words_in_visual}
\end{figure}

\begin{figure}
  \includegraphics[width=\linewidth]{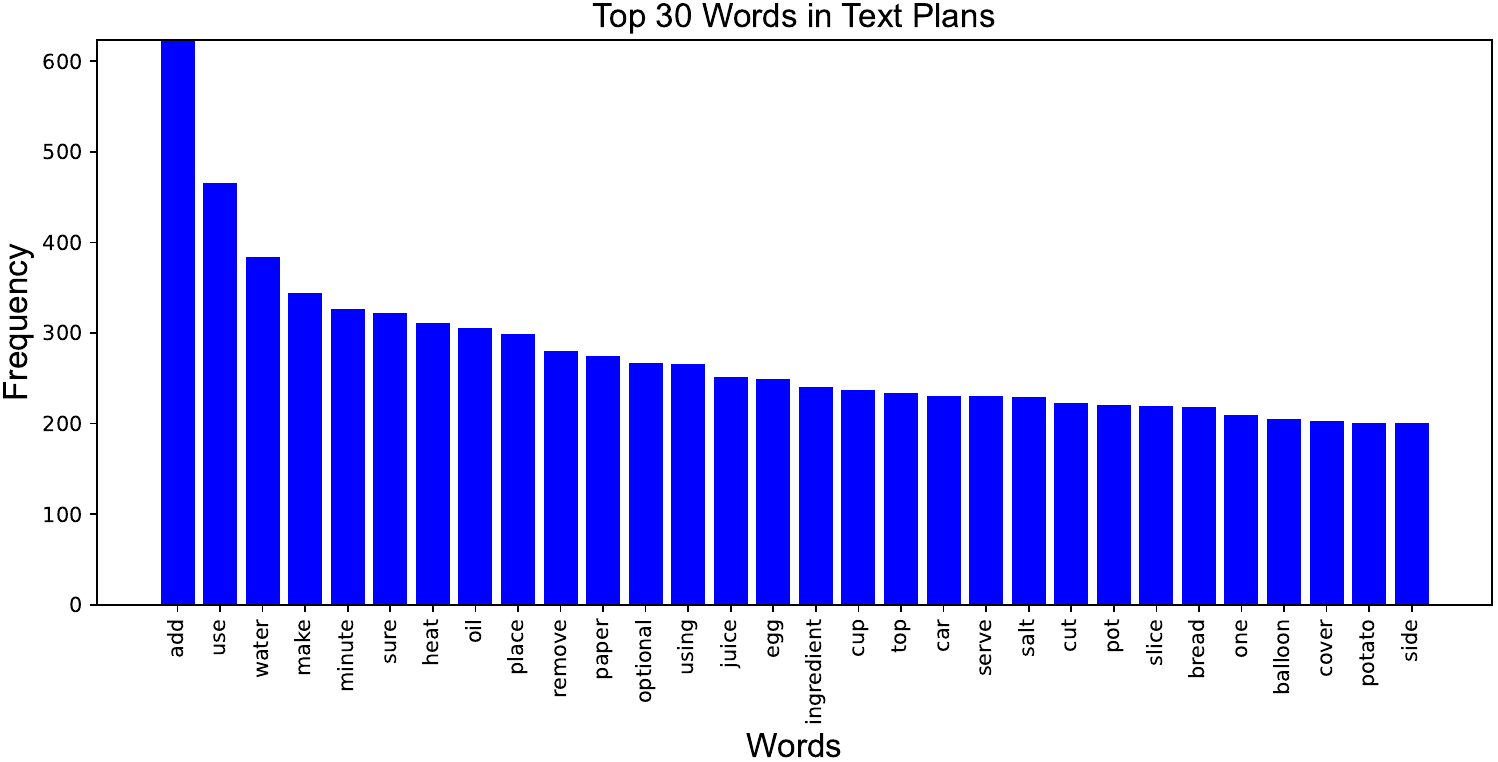}
  \caption{Top 30 Words for Seen Tasks, Vanilla Model}
  \label{fig:vanilla/vanilla_seen_top30words_in_textplans}
\end{figure}

\begin{figure}[htb]
  \includegraphics[width=\linewidth]{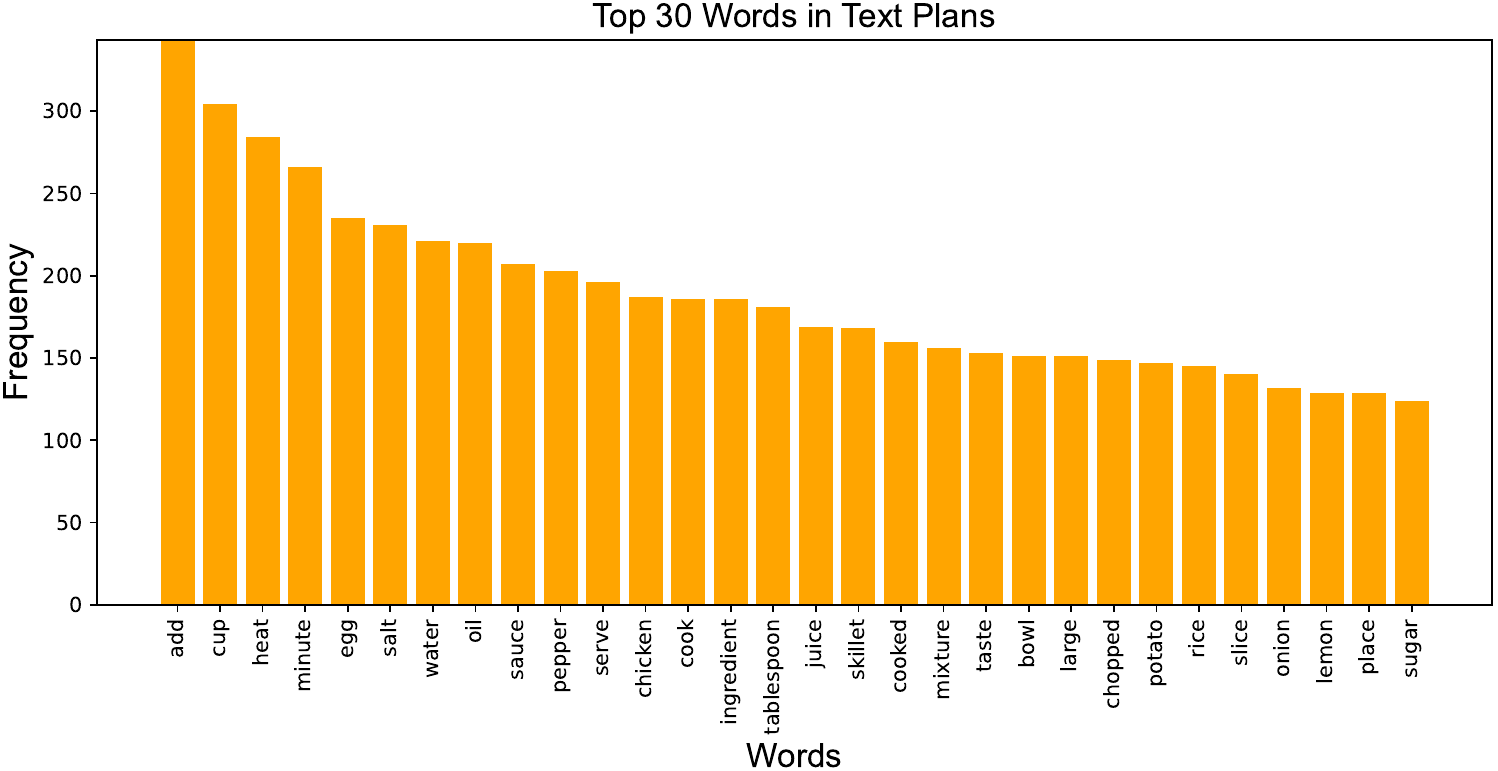}
  \caption{Top 30 Words for Unseen Tasks, Vanilla Model}
  \label{fig:vanilla/vanilla_unseen_top30words_in_textplans}
\end{figure}

\begin{figure}[htb]
  \includegraphics[width=\linewidth]{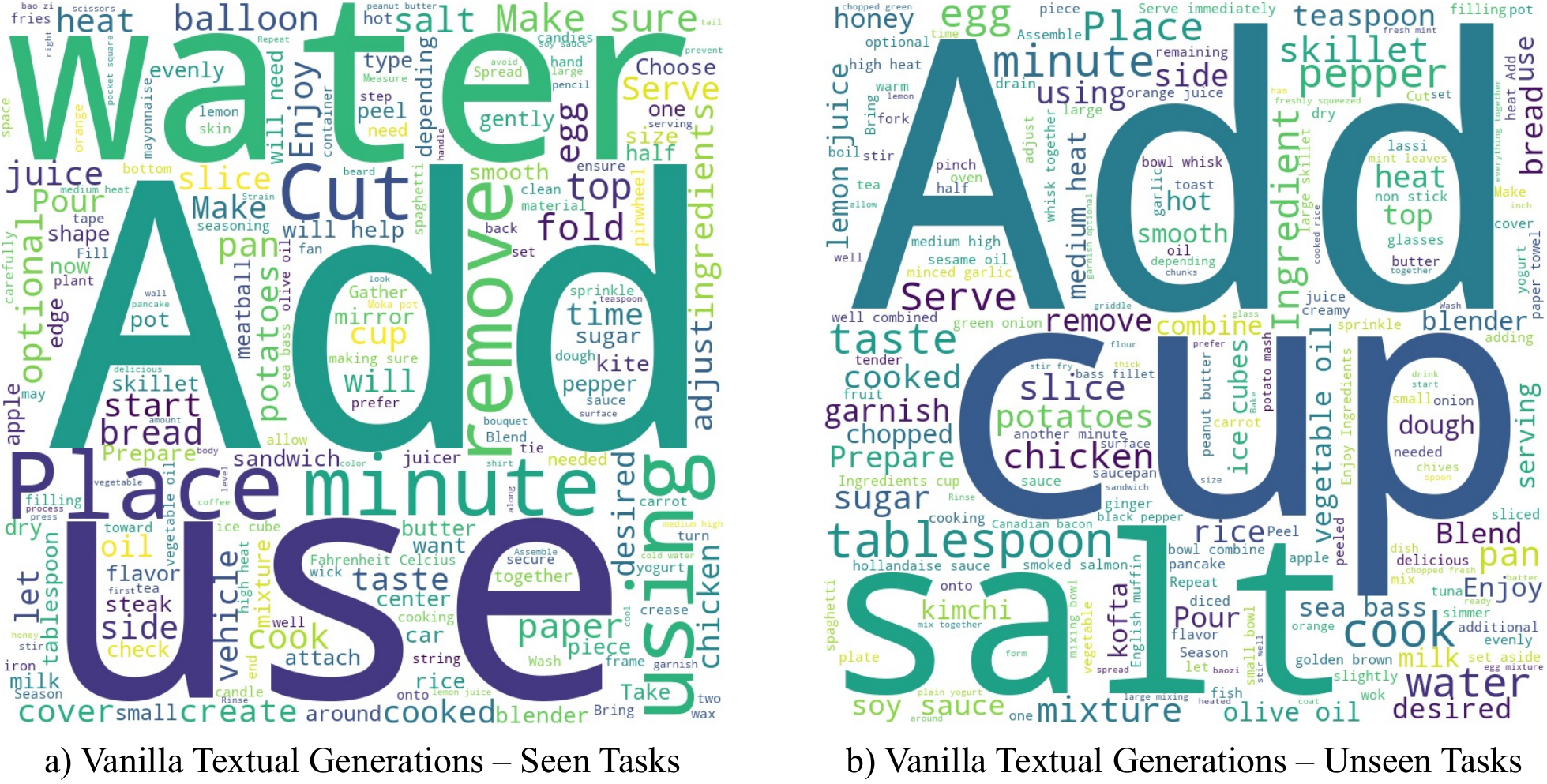}
  \caption{Word cloud distributions of the vanilla textual generations for seen and unseen tasks in the Daily-PP.}
  \label{fig:wdc_vanilla}
\end{figure}

\begin{figure}[!htb]
  \includegraphics[width=\linewidth]{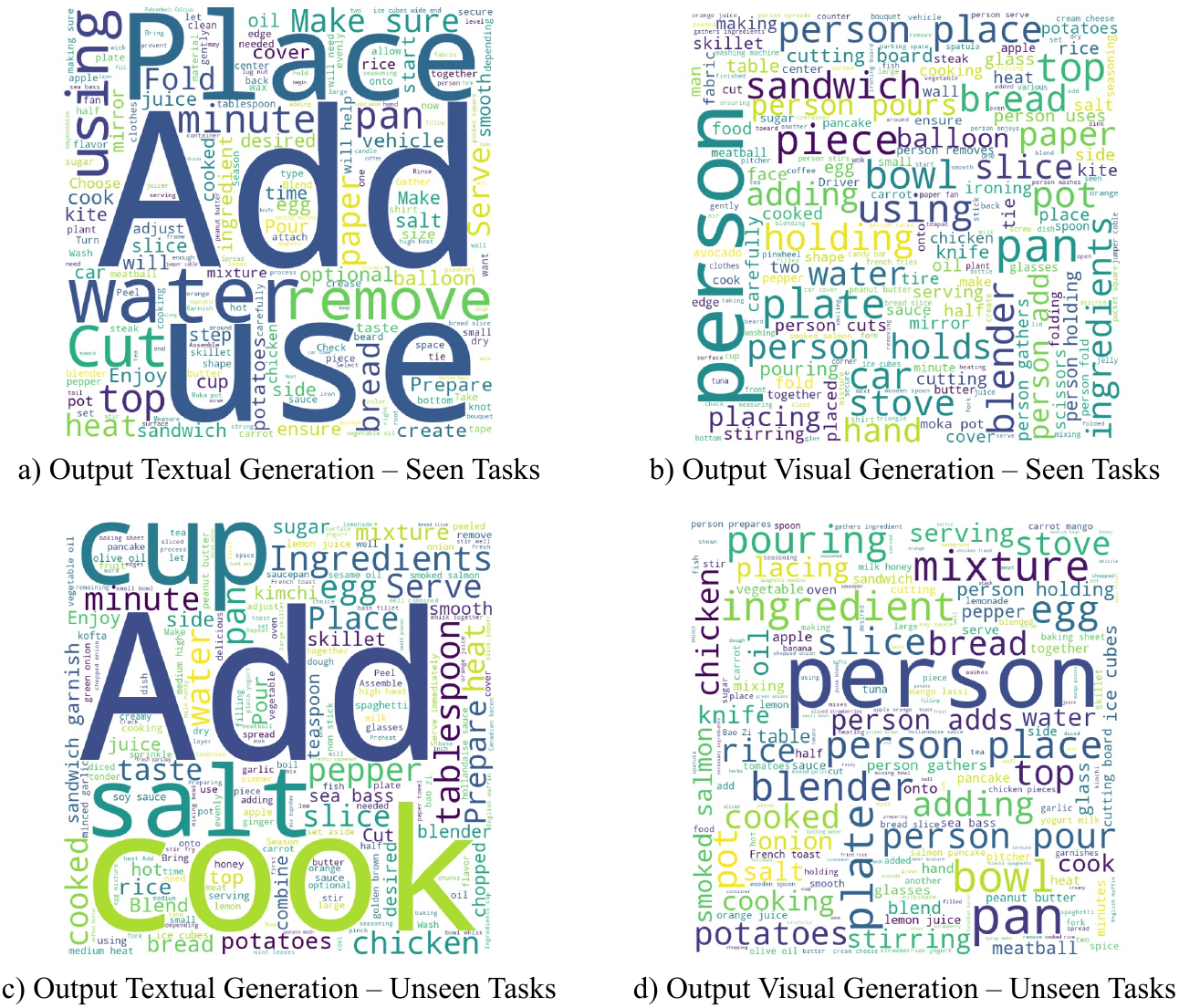}
  \caption{Word cloud distributions for VG-TVP's textual generations for seen and unseen tasks in the Daily-PP.}
  \label{fig:wdc_output}
\end{figure}

In seen tasks, the most frequently generated words by VG-TVP were "add \& ingredient" for Text, "use \& add" for Context, and "car \& pan" for Visual; "add \& use" by Baseline. In unseen tasks, VG-TVP's most frequently generated words were "ingredient \& add" for Text, "cup \& add" for Context, and "pan \& bowl" for Visual; "add \& cup" by Baseline.

\subsection{The Impact of V2T-B and FoC}
V2T-B is the method that textualizes the scenes of IVs by using a video captioning algorithm. VLog is used to generate captions from IVs. We focus on visual information rather than audio information because of the scene audio conflicts, and lack of audio in some videos. This alignment problem between visual and audio is solved via zero-shot prompting.

In our methodology, baseline videos are generated based on the textual context derived from vanilla textual instructions. Vanilla textual baselines have shown notable competence in delivering verbal instructions for a variety of tasks. Yet, when it comes to transferring these textual instructions into video format, their performance in terms of efficiency and visual effectiveness may falter. We propose integrating the FoC via IVs could substantially improve MPP. Figure~\ref{fig:FoCw1} shows an example of the FoC effect. Both figures are generated according to the \textit{Text: "Strain and Serve"}, and \textit{Context: "Once the tea has steeped, it is time to strain the leaves and serve the tea. You can use a strainer or infuser to remove the tea leaves from the water. Once the tea is strained, you can pour it into cups or mugs and add any desired sweeteners or flavorings.}. However, the FOC leverages the text and context to generate a special text as a Visual for output video generation. For this example, the VG-TVP generates a Visual: "The person pours the brewed tea into cups or a serving pitcher and enjoys.". That provides more accurate, text-aligned video plans (Figure~\ref{fig:Qualitative_Szechuan}) that consist of augmented action and state information for the high-level MPP tasks.

FoC reorders and aligns all the steps captured from IVs for the relevant task (Figure \ref{fig:alignment_FoC}). There might be a few instances where FoC, due to incorrect information captured by VLog in the IVs (e.g., an apple being mislabeled as an orange or lemon based on its color), may also misorder information. In such cases, VG-TVP addresses this challenge by leveraging the vanilla text and aligning with it while generating visually grounded text plans. This alignment is crucial for providing relevant visual information to VG-TVP, enabling the generation of augmented MPP content.

\begin{figure}[htb]
\begin{center}
    \includegraphics[width=\linewidth]{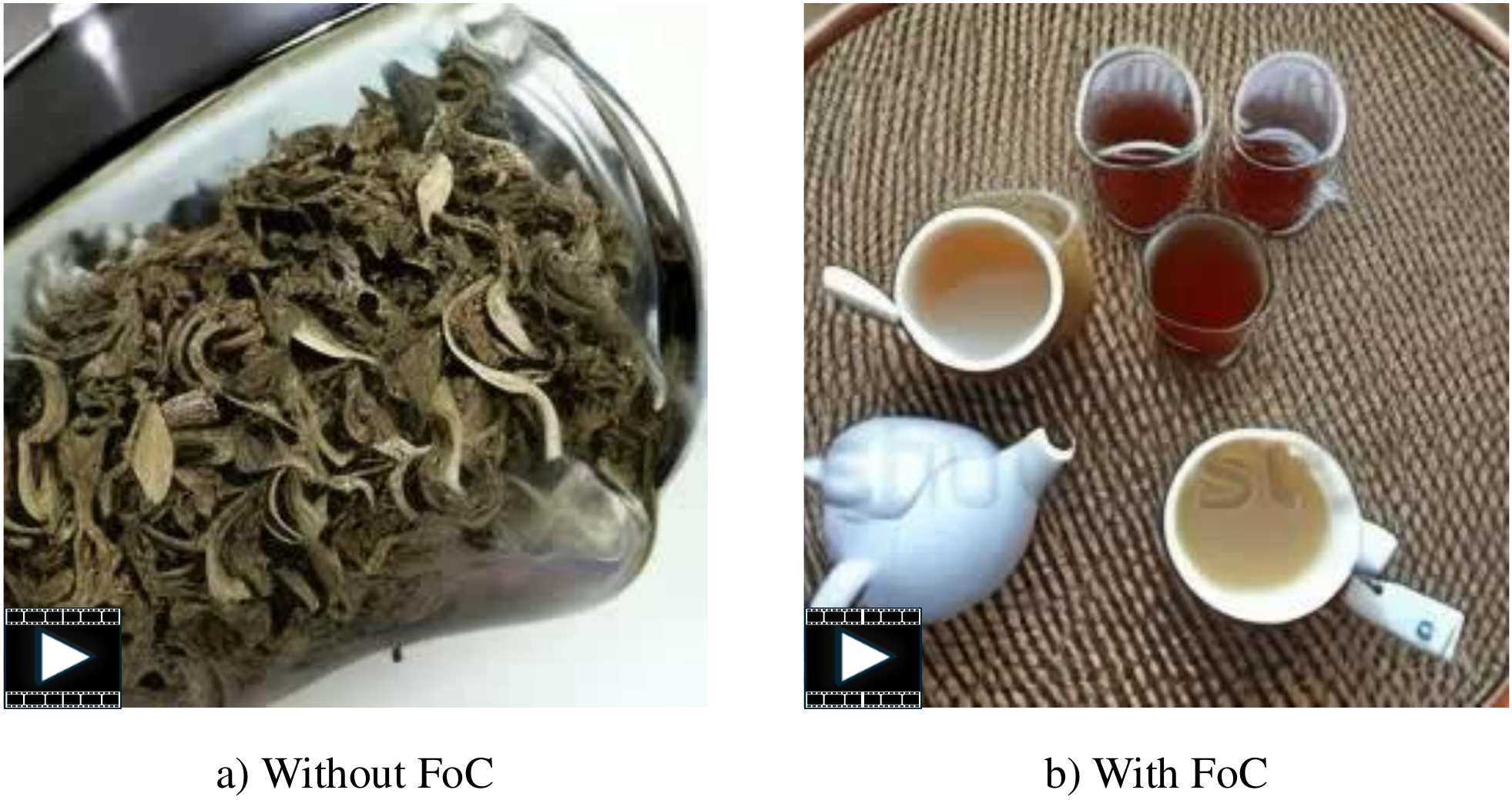}
\end{center}
  \caption{Impact of FoC for "How to Brew a Pot of Tea?".}
  \label{fig:FoCw1}
\end{figure}

\subsection{Human Evaluation Survey}

We use the T2V model for 2 reasons. (1) Compatibility of the models: ModelScope was chosen due to its superior realistic video quality. (2) Testing: We utilized 7 LLMs to generate over 6000 videos, evaluated by human subjects under consistent protocols. The human evaluation metric was realized based on "textual and visual informativeness, temporal coherence, and plan accuracy". A total of 28 participants were involved in the survey. No subject was shown the same task twice. If a subject has already seen \textit{"Kimchi Fried Rice"} task generated by the LLama2-7B-q4 model, they would not be shown a different version of the same task generated by another model. Each participant was over 21 years old and possessed fluency in English. They were only asked to compare two different sequences from the perspective of these four aspects. An example of the survey is shown in Figure~\ref{fig:Survey}. The results derived from the analysis of the collected data are presented in Tables 1 and 2 in the main paper.

\subsection{Qualitative Comparison Example}
This part consists of two different examples: 1) $2$ different visualization prompts' text-video plans versus VG-TVP's text-video plans. 2) The comparison of GPT-3.5's text-video plans and VG-TVP's text-video plans.

\subsubsection{Different Prompting with Visualized Instructions.} LLMs can generate text plans for tasks they have never encountered before, leveraging their zero-shot reasoning capabilities. However, this might cause issues of hallucination. Therefore, it is pertinent to investigate whether alternative prompt engineering methods, beyond the vanilla approach, can enhance the performance of LLMs in generating accurate procedural plans. This example aims to compare the performance of these alternative prompts against the proposed VG-TVP framework. 

We explored $2$ different prompts to generate text and video plans for the "How to make pancakes?" example. First, we apply \textit{"What is the step-by-step procedure for how to make pancakes? Rewrite the textual instruction of making pancakes with visualized instruction pair-wisely in a template $<$text$>$ $<$context$>$, and $<$visual$>$ separately."} as $f_{prompt}(alignment)$. Secondly, we apply \textit{"What is the step-by-step procedure for how to make pancakes? Rewrite the textual instruction of pancake with visualized video instruction pair-wisely in a template $<$text$>$ $<$context$>$, and $<$visual$>$ separately."} as $f_{prompt}(alignment)$. Then we use their text plans to generate video plans. Finally, we apply the VG-TVP model to generate the MPP of the relevant task. The final text and video plans and their comparisons are shown in Figures~\ref{fig:prompt1},~\ref{fig:prompt2}, and~\ref{fig:prompt_vgtvp}. The final text and video plan results show the superiority of VG-TVP.

\subsubsection{GPT-3.5 vs. VG-TVP.} In our use case, we requested human subjects to evaluate the generated textual and video generations (relevant steps) in terms of temporal coherence and plan accuracy. While the generated visuals support our hypothesis by being more successful and human-centred, GPT3.5's textual generations tend to reduce the overall number of steps required to complete the task. Although the exact reason for this behaviour is difficult to determine, given the closed nature of GPT3.5 and the unknown data used during its training process, it might stem from the model's reliance on exact visual matches between vanilla textual and FoC outputs. If an exact match is not provided between them, GPT3.5 opts to use only the matching visuals for text, context, and visual generation. This results in a reduced number of overall steps, leading to deficiencies in temporal coherence and plan accuracy. Consequently, the observed lower performance in temporal coherence and plan accuracy (shown in Table 2) is not due to a negative enhancement from the visual injections but rather due to the expectations of exact visual matches. When considering other performance evaluations, we can also identify positive enhancements for GPT3.5. Therefore, we would like to explain it with an example in different aspects. Textuals generated in 2 different stages are shown in Figure~\ref{fig:GPTExample} below. (1) Vanilla Textual by GPT3.5 and (2) VG-TVP Generated Textuals.

\begin{itemize}
\item \textbf{Visual Description:} The visual description helps users by visualizing the task to make it more intuitive. For example, the visual texts of Step 10 and Step 11 of VG-TVP in Figure~\ref{fig:GPTExample} show the final version of the product.
\item \textbf{Alignment of Text-Video Pairs:} It represents the importance of aligned text-video pairs which provide a comprehensive guide, aiding users in visualizing each step.
\item \textbf{Focused on Key Actions:} It provides avoiding unnecessary information and ensuring clarity. While Step 8 of GPT3-5 (Figure~\ref{fig:GPTExample}) explains the step ignoring the human effect, VG-TVP highlights that effect. VG-TVP's Step 8 is more human-centered, acknowledging that factors like stove structure and heat power can influence cooking.
\item \textbf{Clear and Concise Steps:} VG-TVP separates the text, context, and visual texts. Therefore, that makes VG-TVP more concise and understanding than baselines. The clear and concise texts provide a detailed, explanatory, and user-friendly guide, ensuring users benefit from outputs.
\end{itemize}

We design another experiment with CLIP scores in different frame rates (5,10, 15, and 20) to highlight the impact of generating videos instead of single static images. That experiment's results show the average of mean similarity scores (MSS) for seen and unseen tasks of baselines, VG-TVP; and also benchmark comparisons.

Results display that VG-TVP consistently outperforms baselines and TIP, highlighting the effectiveness of our approach across various task types and frame rates.

For SEEN tasks (Table~\ref{table:seen_task_mss}), VG-TVP achieves higher MSS than the baselines at every frame rate, with an improvement margin that increases as the frame rate decreases. For instance, at a frame rate of 20, VG-TVP achieves an average of MSS 0.3189, compared to the baseline's 0.2946. This margin further widens at a frame rate of 5, where VG-TVP scores 0.3216, significantly outperforming the baseline's 0.2969. These results illustrate VG-TVP's robustness in generating coherent, high-similarity outputs even at lower frame rates, which is crucial for procedural tasks where step-by-step accuracy is the highest.

In UNSEEN tasks (Table~\ref{table:unseen_task_mss}), VG-TVP also surpasses baseline models with similar trends across frame rates. For example, at a frame rate of 20, VG-TVP achieves an average of MSS 0.3268 compared to the baseline's 0.3009, and at a frame rate of 5, VG-TVP reaches 0.3302 versus the baseline's 0.3027. When benchmarked against TIP, VG-TVP's performance is further validated, achieving an average MSS of 0.3290, substantially higher than TIP's 0.3025. This consistent improvement across SEEN and UNSEEN tasks and in comparison, with TIP underscores VG-TVP's superiority in generating visually and temporally coherent outputs, reinforcing its suitability for practical applications requiring detailed, dynamic multimodal procedural plans.

\begin{table}
\begin{center}
\begin{tabular}{ccccc}
\hline
Frame Rate  & \multicolumn{2}{c}{SEEN Tasks (Idea 1)}   &    \\
            & Baselines                                 & VG-TVP                                  \\
\hline
20          & 0.294634078                              & 0.318871641                              \\
15          & 0.296215234                              & 0.320155963                              \\
10          & 0.296341701                              & 0.320829739                              \\
5           & 0.296913945                              & 0.321579028                              \\
\hline
\end{tabular}
\end{center}
\caption{Average MSS of Baselines and VG-TVP for Seen Tasks (Idea 1).}
\label{table:seen_task_mss}
\end{table}

\begin{table}
\begin{center}
\begin{tabular}{ccccc}
\hline
Frame Rate  & \multicolumn{2}{c}{UNSEEN Tasks (Idea 2)}   &    \\
            & Baselines                                   & VG-TVP                            \\
\hline
20          & 0.3009426                              & 0.3268314                              \\
15          & 0.3020536                              & 0.3290246                              \\
10          & 0.3025034                              & 0.3290118                              \\
5           & 0.3027169                              & 0.3301511                              \\
\hline
\end{tabular}
\end{center}
\caption{Average MSS of Baselines and VG-TVP for Unseen Tasks (Idea 2).}
\label{table:unseen_task_mss}
\end{table}

\subsection{LLM Evaluation Protocol with Socratic Method}

The Socratic Model represents the resolution of complex tasks through a series of questions. It involves various principles (Chang 2023), such as identifying key points of the tasks, proposing cause-and-effect relationships for tasks, presenting text plan examples, and asking evaluation aspects questions. Thus, we leverage each of these principles using ChatGPT-4o. The evaluation protocol evaluates baselines and VG-TVP on $4$ aspects as in the human evaluation metric. There are 5 and 3 randomly chosen tasks from each domain that were evaluated for seen (Candy Bouquet, Change a Tire, Kimchi Fried Rice, Moka Pot Coffee, and Pancake) and unseen (Chicken Fried Rice, Carrot Mango Lassi and Egg Benedict) tasks, respectively. This evaluation considers the semantic relevance of the task plans generated by baselines and VG-TVP. Each scored out of $25$ points, thereby out of $100$ points in total. All results for the seen and unseen tasks are shown in Table~\ref{table:LLM_seen_task} and Table~\ref{table:LLM_unseen_task} in terms of Textual Informativeness, Visual Informativeness, Temporal Alignment, and Plan Accuracy which are displayed as T.I., V.I., T.A., and P.A., respectively. The final protocol is as follows:

\subsubsection{Textual Informativeness. (25 points)} This aspect assesses the instructional text's clarity, comprehensiveness, detail, and overall quality. It consists of \textit{Comprehensiveness (10 points), Clarity\&Precision (10 points), and Detail Level (5 points)}. \textit{"Comprehensiveness"} covers essential steps (5 points) and additional information (5 points). Essential steps evaluate whether all necessary steps are included. Additional information considers the inclusion of supplementary details, such as background information, optional steps, or alternative methods. \textit{"Clarity\&Precision"} covers language clarity (5 points) which assesses the language's clarity and simplicity, and specificity (5 points) which evaluates the precision of instructions, including clear descriptions of actions, measurements, and conditions. Finally, the \textit{"Detail Level"} covers ingredient\&tool specifications (2.5 points) and step-by-step breakdown (2.5 points). Ingredient\&tool specifications represent specificity in listing ingredients, tools, and equipment. Step-by-step breakdown ensures complex actions are broken down into manageable components.

\subsubsection{Visual Informativeness. (25 points)} This measures the effectiveness of visual descriptions in conveying the steps and elements. It consists \textit{Visualization Cues (10 points), Imagery Description (10 points), and Use of Examples and Analogies (5 points)}. \textit{Visualization Cues} evaluate descriptive elements that help visualize actions and objects. \textit{Imagery and Descriptions} assess how well the instructions create a visual of the process. \textit{Use of Examples and Analogies} consider examples or illustrations that aid in visualization.

\subsubsection{Temporal Alignment. (25 points)} This evaluates the logical sequencing and timing information provided in the instructions. It consists \textit{Chronological Order (10 points), Time Indications (10 points), and Simultaneous Actions (5 points)}. \textit{Chronological Order} checks if the steps are presented in a logical sequence. \textit{Time Indications} evaluate the presence and accuracy of time-related information. \textit{Simultaneous Actions} considers how well the instructions handle simultaneous or overlapping actions.

\subsubsection{Plan Accuracy. (25 points)} This measures the accuracy and practicality of the instructions in guiding the user to successfully complete the task. It consists of \textit{Correctness of Steps (15 points), Consistency (5 points), and Practicality\&Feasibility (5 points)}. \textit{Correctness of Steps} assesses whether the steps align with the actual task requirements. \textit{Consistency} evaluates if the instructions are consistent and free of contradictions. \textit{Practicality\&Feasibility} considers the real-world applicability and ease of execution.

\subsubsection{Scoring and Feedback.} The sum of points from each aspect provides an overall score out of 100. Additionally, detailed feedbacks for each aspect, are provided to highlight strengths and areas for improvement. For example, (in terms of textual informativeness), LLM (ChatGPT-4o) evaluator generates comments for baselines and VG-TVP (while both using the model GPT-3.5) \textit{"(21.5/23). The instructions provide a clear outline of the necessary steps and ingredients, including optional adjustments. However, they lack detailed explanations and context, such as the reasoning behind ingredient choices."}, and \textit{"(23.5/25). The instructions are highly detailed, clear, and comprehensive, covering all necessary components. They include specific measurements and thorough explanations, making the process easy to follow."} respectively. Therefore, for fair benchmarking, we compare the baseline's text plans and VG-TVP's text plans while they use the same model (GPT-3.5 for this example).

\begin{table}
\begin{center}

\begin{tabular}{lccccc}
\hline
Models        & T.I. & V.I. & T.A. & P.A. & Total \\
\hline
Llama2-7B-q4  &                         &                        &                    &               &        \\
Baseline      & 20.6                    & 7.5                    & 18                 & 18.9          & 65     \\
VG-TVP        & 23.6                    & 22.7                   & 22                 & 23            & 91.3   \\
\hline
Llama2-7B-q8  &                         &                        &                    &               &        \\
Baseline      & 20.8                    & 7.4                    & 18.4               & 19            & 65.6   \\
VG-TVP        & 23.7                    & 22.2                   & 21.8               & 22.8          & 90.5   \\
\hline
Llama2-13B-q4 &                         &                        &                    &               &        \\
Baseline      & 21.2                    & 8.3                    & 18.8               & 19.5          & 67.8   \\
VG-TVP        & 24.1                    & 23.2                   & 22                 & 23.2          & 92.5   \\
\hline
Llama2-13B-q8 &                         &                        &                    &               &        \\
Baseline      & 21.3                    & 8.1                    & 18.7               & 19.3          & 67.4   \\
VG-TVP        & 24                      & 22.6                   & 22.2               & 23.5          & 92.3   \\
\hline
Mistral-7B-q4 &                         &                        &                    &               &        \\
Baseline      & 20.4                    & 7.6                    & 18.2               & 19.3          & 65.5   \\
VG-TVP        & 23.7                    & 22.6                   & 21.9               & 22.9          & 91.1   \\
\hline
Mistral-7B-q8 &                         &                        &                    &               &        \\
Baseline      & 20.9                    & 7.9                    & 18.3               & 19.7          & 66.8   \\
VG-TVP        & 23.8                    & 22.6                   & 22.2               & 23            & 91.6   \\
\hline
ChatGPT3.5    &                         &                        &                    &               &        \\
Baseline      & 22.2                    & 8.6                    & 19                 & 19.8          & 69.6   \\
VG-TVP        & 24                      & 22.6                   & 21.9               & 23.6          & 92.1   \\
\hline
\end{tabular}
\end{center}
\caption{Average results of LLM evaluations (chosen from $5$ random tasks, $1$ from each domain) for SEEN tasks. ChatGPT-4o compares baselines and VG-TVP textual.}
\label{table:LLM_seen_task}
\end{table}

\begin{table}
\begin{center}

\begin{tabular}{lccccc}
\hline
Models        & T.I. & V.I. & T.A. & P.A. & Total \\
\hline
Llama2-7B-q4  &                         &                        &                    &               &        \\
Baseline      & 21.3                    & 9.8                    & 18.7               & 19.3          & 69.2   \\
VG-TVP        & 23.0                    & 22.0                   & 21.7               & 23.0          & 89.7   \\
\hline
Llama2-7B-q8  &                         &                        &                    &               &        \\
Baseline      & 21.0                    & 10.2                   & 18.3               & 19.3          & 68.8   \\
VG-TVP        & 23.2                    & 22.5                   & 22.0               & 22.7          & 90.3   \\
\hline
Llama2-13B-q4 &                         &                        &                    &               &        \\
Baseline      & 21.7                    & 11.0                   & 19.2               & 19.8          & 71.7   \\
VG-TVP        & 23.2                    & 22.5                   & 22.0               & 23.0          & 90.7   \\
\hline
Llama2-13B-q8 &                         &                        &                    &               &        \\
Baseline      & 21.5                    & 10.7                   & 18.8               & 20.3          & 71.3   \\
VG-TVP        & 23.5                    & 22.5                   & 22.0               & 23.7          & 91.7   \\
\hline
Mistral-7B-q4 &                         &                        &                    &               &        \\
Baseline      & 21.5                    & 12.2                   & 18.7               & 20.0          & 72.3   \\
VG-TVP        & 23.8                    & 23.0                   & 22.3               & 23.2          & 92.3   \\
\hline
Mistral-7B-q8 &                         &                        &                    &               &        \\
Baseline      & 22.3                    & 13.0                   & 19.5               & 20.5          & 75.3   \\
VG-TVP        & 24.0                    & 23.0                   & 22.3               & 23.3          & 92.7   \\
\hline
ChatGPT3.5    &                         &                        &                    &               &        \\
Baseline      & 22.5                    & 11.0                   & 19.8               & 20.8          & 74.2   \\
VG-TVP        & 24.0                    & 22.8                   & 22.5               & 23.7          & 93.0   \\
\hline
\end{tabular}
\end{center}
\caption{Average results of LLM evaluations (chosen from $3$ random tasks, $1$ from each domain) for UNSEEN tasks. ChatGPT-4o compares baselines and VG-TVP textual.}
\label{table:LLM_unseen_task}
\end{table}

\begin{figure*}[htb]
\begin{center}
\includegraphics[width=\linewidth]{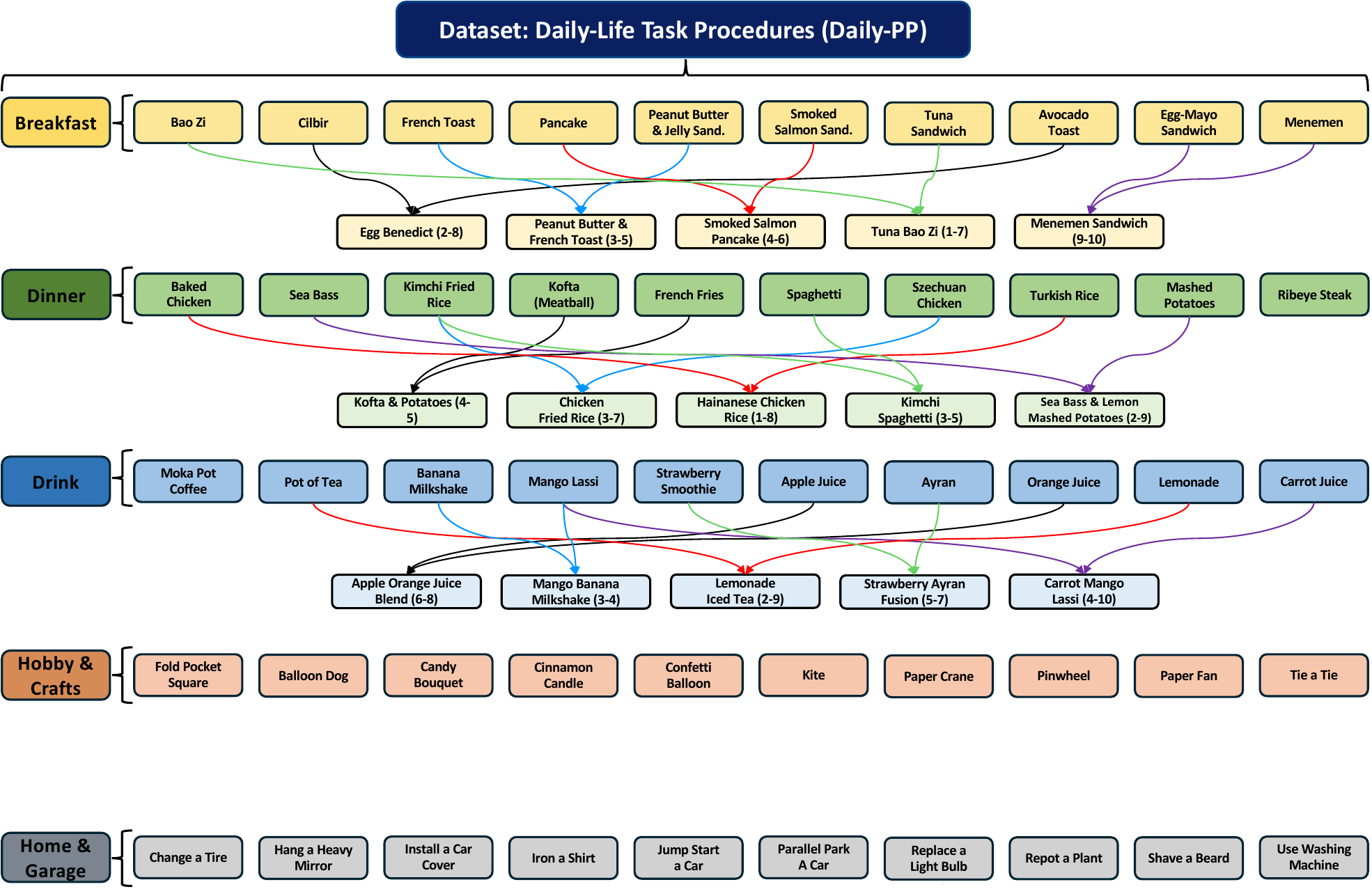}
\end{center}
   \caption{Daily-PP: Daily Life Task Procedures Dataset Structure}
\label{fig:Dataset}
\end{figure*}

\begin{figure*}[htb]
\begin{center}
\includegraphics[width=\linewidth]{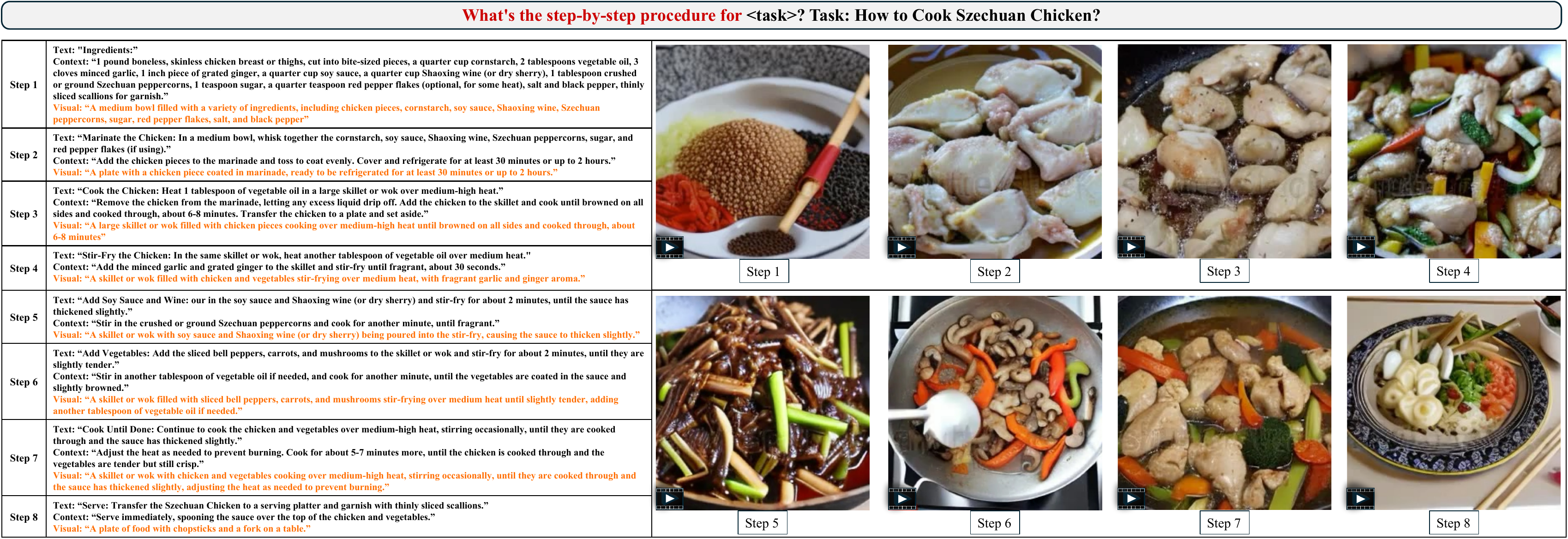}
\end{center}
   \caption{Qualitative result of VG-T2V Model (Ours) for the "How to Cook Szechuan Chicken?" task. Visual texts are used to generate task videos for the VG-TVP model.}
\label{fig:Qualitative_Szechuan}
\end{figure*}

\begin{figure*}[htb]
\begin{center}
\includegraphics[width=\linewidth]{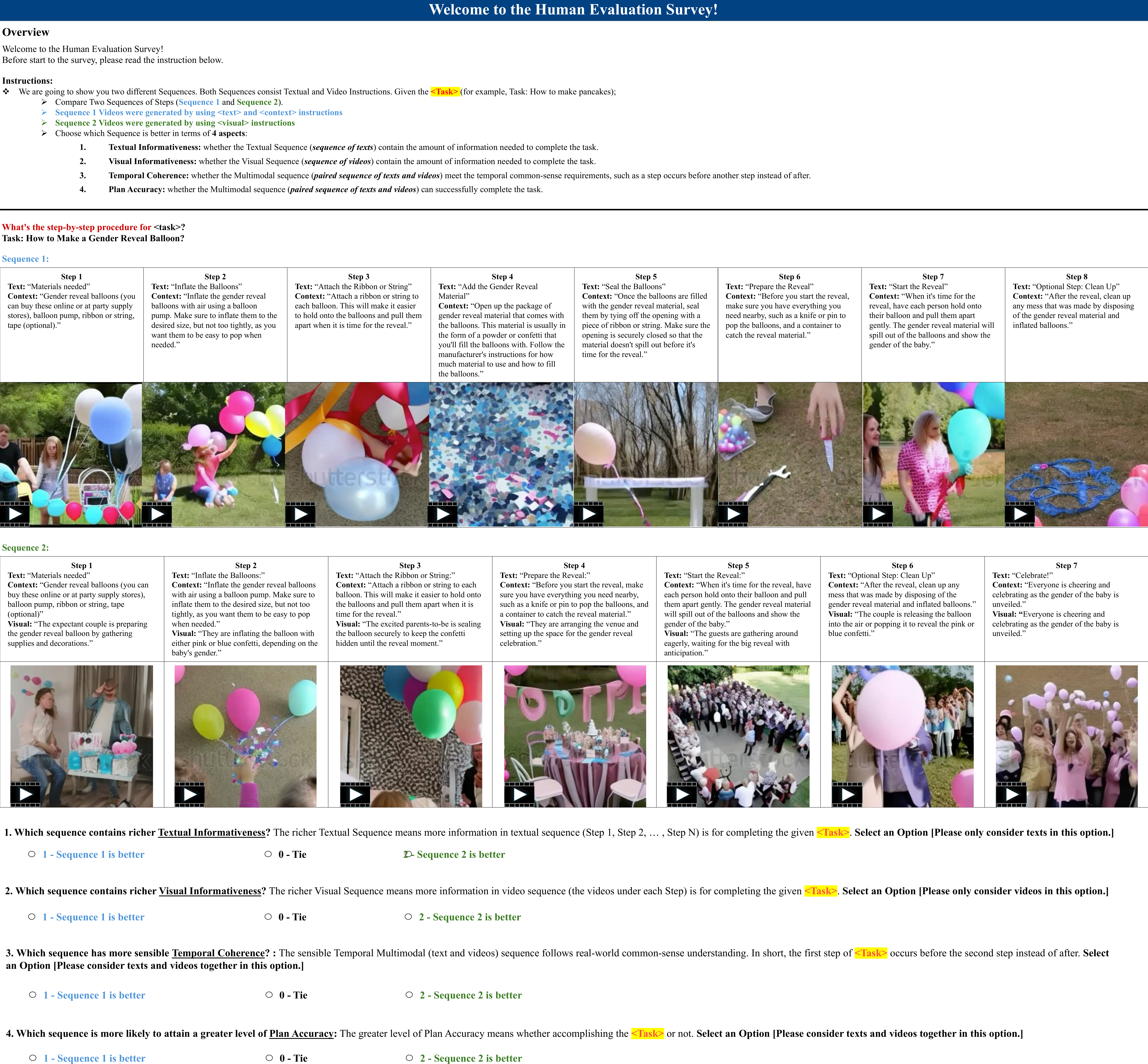}
\end{center}
   \caption{The survey example for the human evaluation study.}
\label{fig:Survey}
\end{figure*}

\begin{figure*}[htb]
\begin{center}
\includegraphics[width=\linewidth]{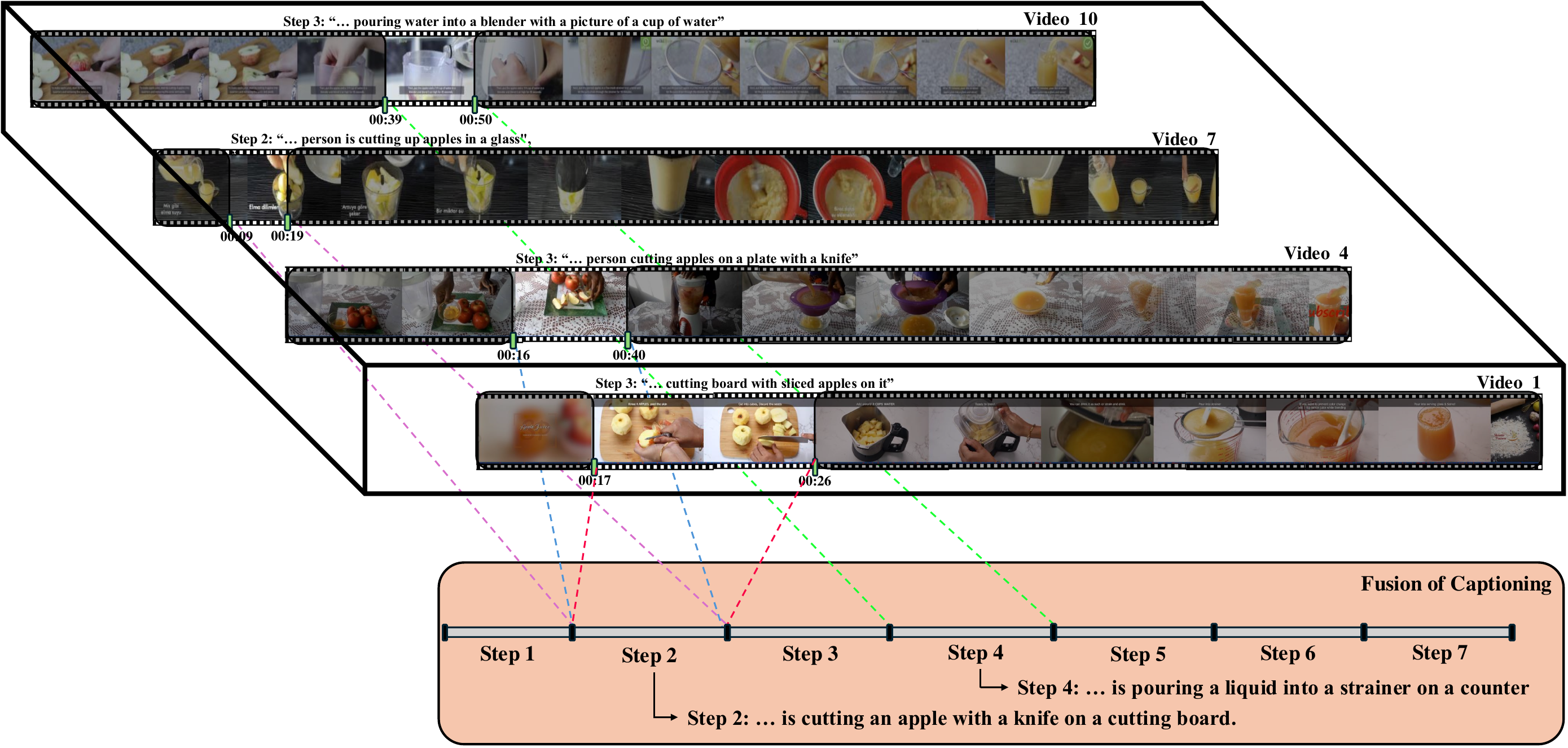}
\end{center}
   \caption{The fundamental representation of the FoC to show how different mismatched steps in IVs are aligned and prepared for generating an accurate MPP content.}
\label{fig:alignment_FoC}
\end{figure*}

\begin{figure*}[htb]
\begin{center}
\includegraphics[width=\linewidth]{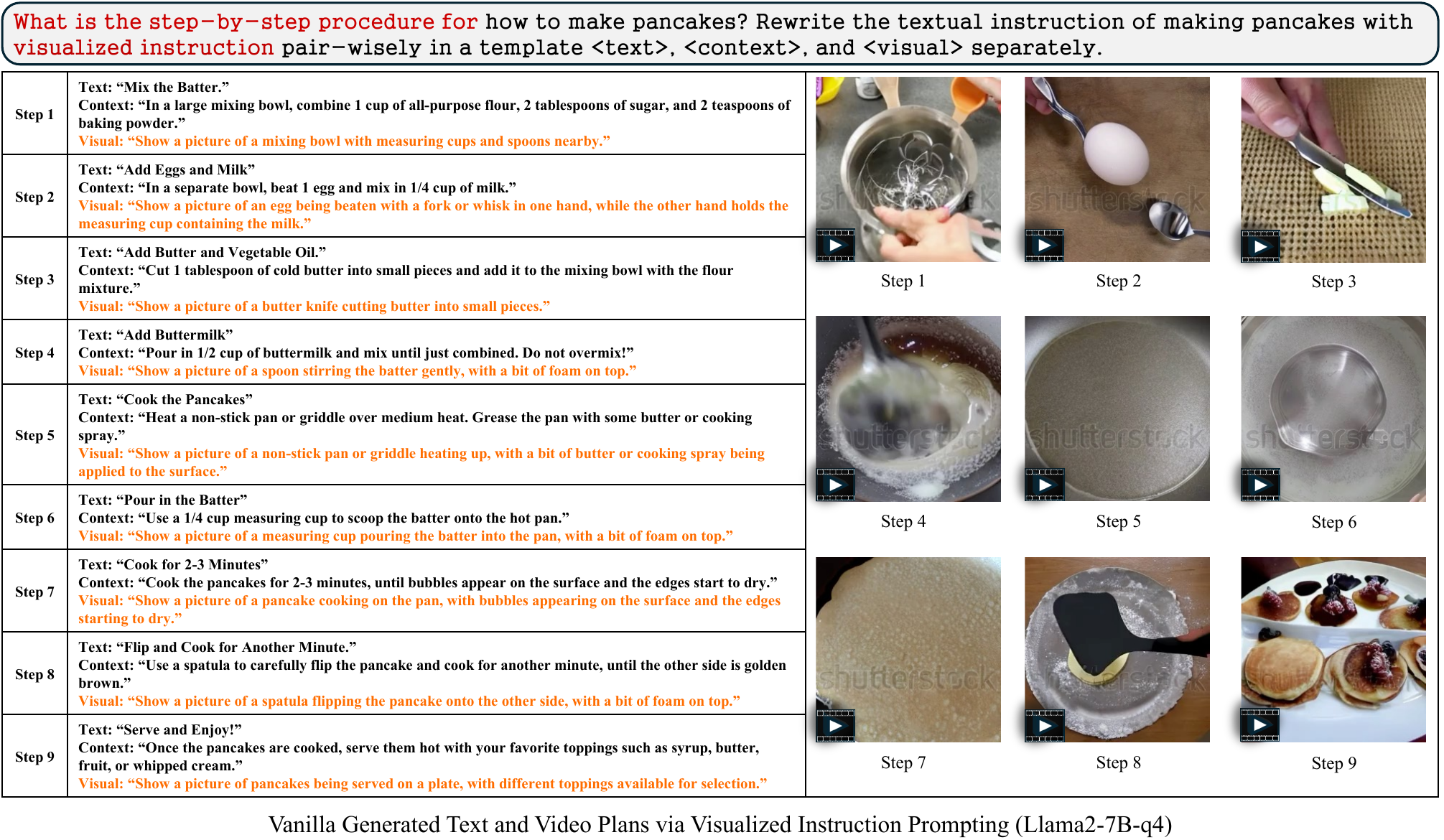}
\end{center}
   \caption{Qualitative Comparison Example with Visualized Instruction Prompting by Llama2-7B-q4, (Task: How to make pancakes?)}
\label{fig:prompt1}
\end{figure*}

\begin{figure*}[htb]
\begin{center}
\includegraphics[width=\linewidth]{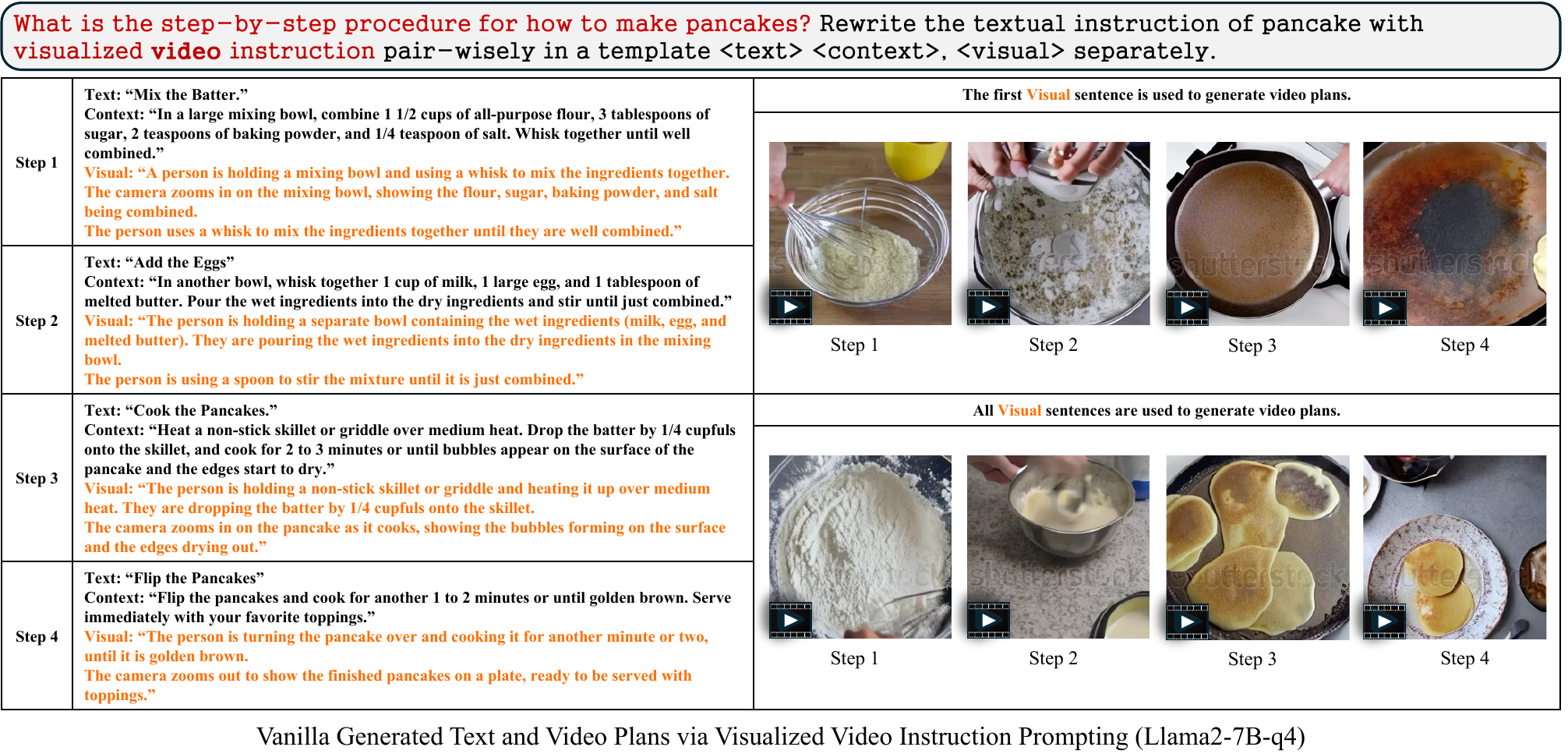}
\end{center}
   \caption{Qualitative Comparison Example with Visualized Video Instruction Prompting by Llama2-7B-q4, (Task: How to make pancakes?)}
\label{fig:prompt2}
\end{figure*}

\begin{figure*}[htb]
\begin{center}
\includegraphics[width=\linewidth]{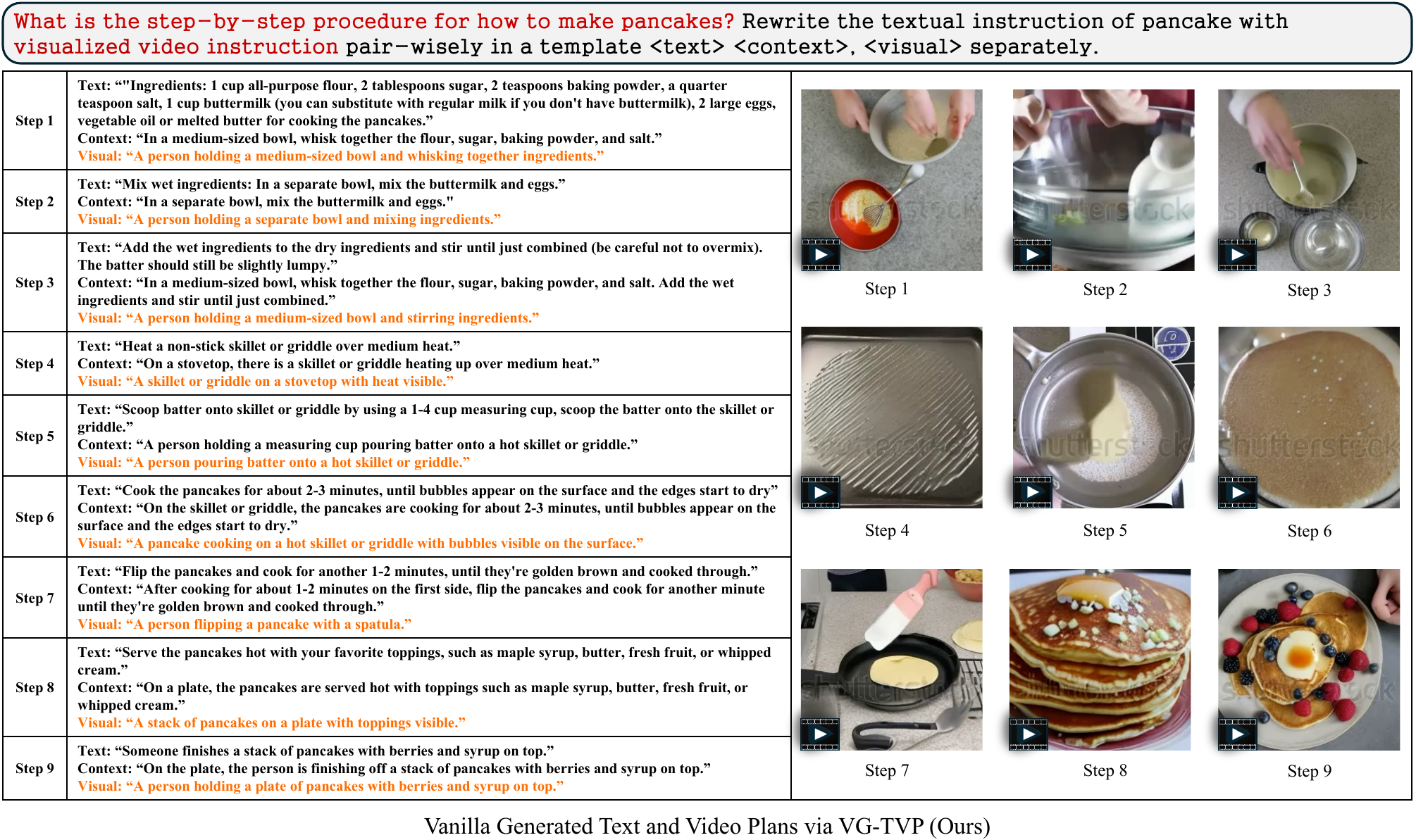}
\end{center}
   \caption{Qualitative Comparison Example by VG-TVP (Ours), (Task: How to make pancakes?)}
\label{fig:prompt_vgtvp}
\end{figure*}

\begin{figure*}[htb]
\begin{center}
\includegraphics[width=0.96 \textwidth, height=0.96 \textheight, keepaspectratio]{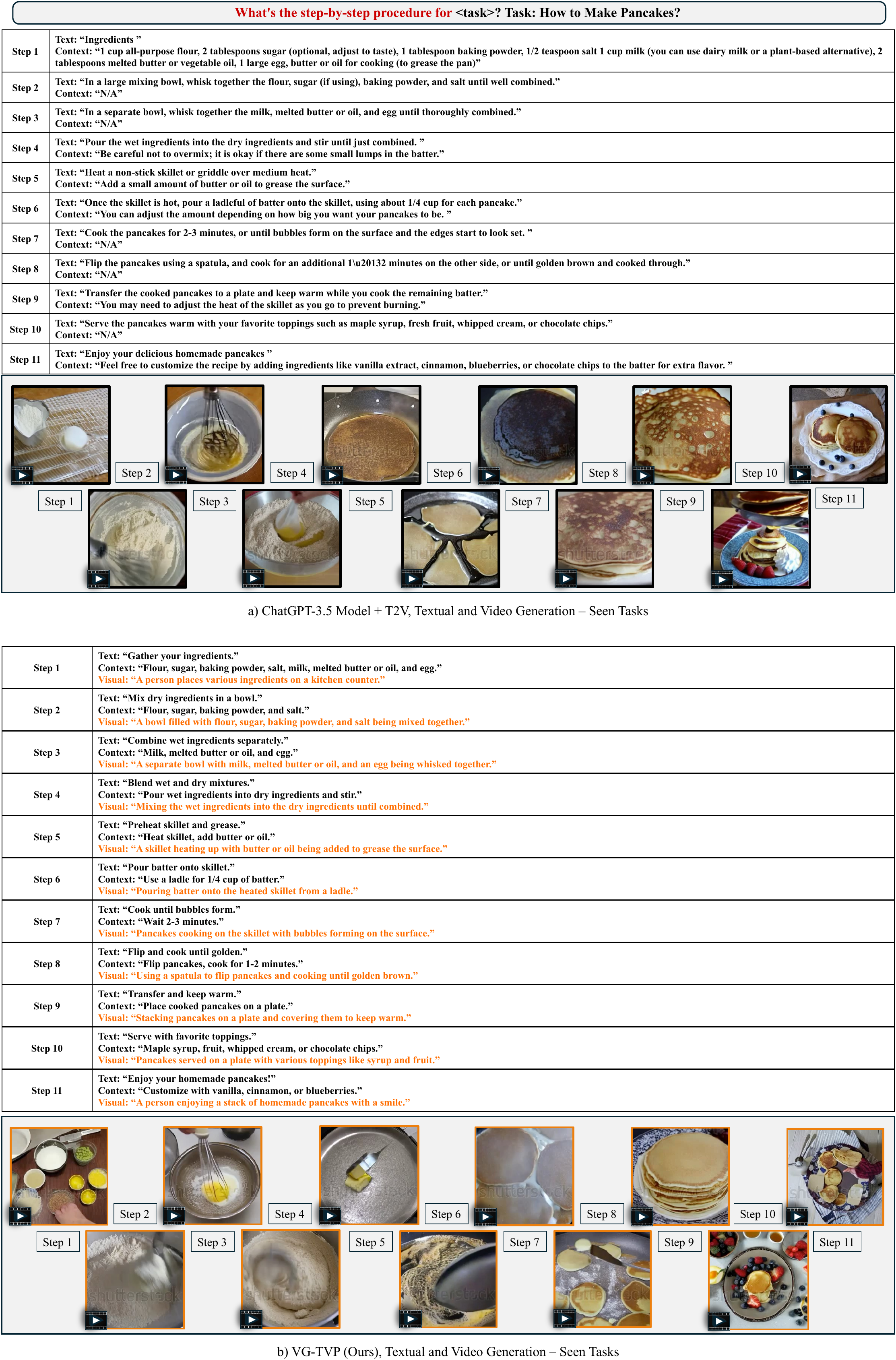}
\end{center}
   \caption{Qualitative Comparison Example: GPT-3.5 vs. VG-TVP. Visuals (orange) are used to generate video plans.}
\label{fig:GPTExample}
\end{figure*}

\end{document}